\documentclass{article}

\PassOptionsToPackage{numbers, sort&compress}{natbib}

\usepackage[preprint]{neurips_2024}

\usepackage[utf8]{inputenc}      
\usepackage[T1]{fontenc}         
\usepackage{url}                 
\usepackage{booktabs}            
\usepackage{amsfonts}            
\usepackage{nicefrac}            
\usepackage{microtype}           
\usepackage[dvipsnames]{xcolor}  
\usepackage{colortbl}            
\usepackage{pifont}              
\usepackage{amssymb}             

\usepackage{epigraph}
\usepackage{amsmath}
\usepackage{enumerate}
\usepackage{multirow}
\usepackage{graphicx}
\usepackage{caption}
\usepackage{subcaption}
\usepackage{siunitx}
\usepackage{physics}
\usepackage[normalem]{ulem}
\usepackage{mathtools}
\usepackage{comment}
\usepackage{tcolorbox}
\usepackage[
  pagebackref,
  breaklinks,
  colorlinks,
  linkcolor=red,
  citecolor=blue,
  filecolor=mydarkblue,
  urlcolor=mydarkblue,
]{hyperref}
\usepackage[capitalise]{cleveref}
\usepackage{colortbl}
\usepackage{enumitem}

\definecolor{mydarkblue}{rgb}{0,0.08,0.8}
\definecolor{lightorange}{rgb}{0.97,0.86,0.80}
\definecolor{modorange}{rgb}{0.9453125, 0.6640625, 0.515625}
\definecolor{modblue}{rgb}{0.37890625, 0.79296875, 0.953125}
\newcommand{\cmark}{\ding{51}}%
\newcommand{\xmark}{\ding{55}}%

\newcommand{\eg}{\textit{e}.\textit{g}.}
\newcommand{\etc}{\textit{etc}}
\let\OLDthebibliography\thebibliography
\renewcommand\thebibliography[1]{
  \OLDthebibliography{#1}
  \setlength{\parskip}{2pt}
  \setlength{\itemsep}{2pt plus 0.3ex}
}

\title{
The Sound of Water: Inferring Physical Properties from Pouring Liquids
}

\author{%
  Piyush Bagad \\
  University of Oxford
  \And
  \hspace{5mm} Makarand Tapaswi \\
  \hspace{5mm} IIIT Hyderabad
  \AND
  Cees G. M. Snoek \\
  University of Amsterdam
  \And
  Andrew Zisserman \\
  University of Oxford
  \\[3mm]
  \hspace{-6cm}\url{https://bpiyush.github.io/pouring-water-website}
}

\AtBeginDocument{\RenewCommandCopy\qty\SI}
\begin{document}

\maketitle

\begin{figure}[h]
  \centering
  \includegraphics[width=\linewidth]{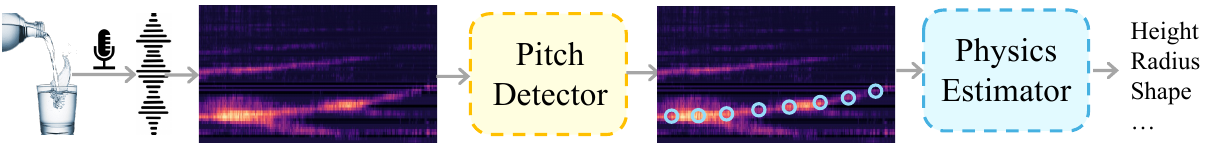} 
  \captionsetup{skip=2mm,font=normal}
  \caption{\textbf{Overview of the problem and approach.} We train a pitch detector without any manual supervision and rely on physics to estimate physical properties merely from the sound of water.}
  \label{fig:my_label}
\end{figure}

\begin{abstract}

We study the connection between audio-visual observations and the underlying physics of a mundane yet intriguing everyday activity: \textit{pouring liquids}. Given only the sound of liquid pouring into a container, our objective is to automatically infer physical properties such as the liquid level, the shape and size of the container, the pouring rate and the time to fill.
To this end, we: (i) show in theory that these properties can be determined from the fundamental frequency (pitch); (ii) train a pitch detection model with supervision from simulated data and visual data with a physics-inspired objective; (iii) introduce a new large dataset of real pouring videos for a systematic study; (iv) show that the trained model can indeed infer these physical properties for real data; and finally, (v) we demonstrate strong generalization to various container shapes, other datasets, and in-the-wild YouTube videos.
Our work presents a keen understanding of a narrow yet rich problem at the intersection of acoustics, physics, and learning. It opens up applications to enhance multisensory perception in robotic pouring.

\end{abstract}

\epigraph{
  \it
  ``The blind man of Puisaux judges of his nearness to the
  fire by the degrees of heat; of the fulness of vessels by
  the sound made by liquids which pours into them; of the
  proximity of bodies by the action of the air on his face.''
}{
  \footnotesize -- Denis Diderot,
  \textnormal{Letter on the Blind (1749)}
}

\section{Introduction}

What can possibly be scientifically interesting about such a mundane chore as pouring a liquid into a glass? We perform this action all the time but barely realise that we effortlessly learn to infer several useful physical properties in the process.
For example, evidence in psychoacoustics suggests that humans can accurately infer the level of the liquid, the time to fill~\cite{Cabe2000HumanST}, the size of the container~\cite{container_size_doi:10.1177/0956797618813319}, and even the temperature of the liquid~\cite{temperature1_https://doi.org/10.1111/joss.12052,temperature2_Agrawal2020HearingWT,tomscott_youtube}, merely from the sound of pouring.
Such inference (\eg, time to fill) allows us to adaptively control our actions (\eg, stopping pouring to prevent spillage) conforming to the \textit{affordance} theory by Gibson~\cite{gibson2014ecological}. In this work, we study the physical phenomenon involved in liquid pouring and explore how it can be used to train machines to infer useful physical properties from sound alone.

Despite its mundaneness, liquid pouring has rich physics underpinning it and has been studied for more than a century~\cite{rayleigh1896theory}.
The crux of this exploration is summarized well by Berg and Stork~\cite{berg1982physics}: ``As the liquid (\eg water) is filled, a sound consisting of an increasing pitch and some (odd) harmonics superimposed with whooshing, gurgling is observed''.
This pitch and the corresponding harmonics are a function of the physical properties. For example, the shape of the pitch depends on the container shape~\cite{radial1-French1983InVV, webster2010use, radial2-PhysRev.13.171}, the range of the pitch depends on the container dimensions~\cite{Cabe2000HumanST}, and the rate change of pitch depends on the pouring flow rate\cite{Cabe2000HumanST}.
Thus, automatically inferring useful physical properties from the sound of pouring necessitates two stages:
(i) detecting the pitch from the raw audio signal, and (ii) recovering these physical properties from the pitch. There are several challenges in training machines to do these purely from sound.

First, such a task requires a fine-grained, time-sensitive understanding of audio while contemporary models focus more on coarse recognition tasks like sound classification~\cite{kay2017kinetics,vggsound,audioset}. Second, the underlying physics of such a niche activity as pouring is not fully developed for a general
container and liquid setup, unlike, say, Newtonian mechanics studied analogously in \cite{phys101}. Third, there is a lack of clean, controlled, and large datasets which are necessary to study physical property estimation by learning.
Fourth, supervision either in the form of pitch annotation or the actual physical properties is difficult to obtain and to use directly in training.

To enable a systematic study, we collect a clean and large dataset of 805 videos of pouring across 50 diverse containers.
For training, we select a subset of containers shaped like cylinders such that we can approximate the underlying physics with that of a cylinder~\cite{Cabe2000HumanST}.
We design an audio network for pitch detection based on \texttt{wav2vec2}~\cite{baevski2020wav2vec} pre-trained on speech data that has characteristic pitch dynamics~\cite{morise2017harvest}. As pitch annotations are hard and ambiguous to obtain at scale, we use supervision from simulated data and visual data. We pre-train the network on simulated sounds of liquid pouring. On real data, we fine-tune the network by visual co-supervision with a physics-inspired objective.
We demonstrate that the co-supervised audio model is able to predict pitch, and hence estimate physical properties, with a performance far exceeding that of multiple previous methods.

Why is this important, though? First, as far as we know, this is the first work to demonstrate human-like capabilities (or better) in predicting physical properties from sound alone -- in fact, we achieve an accuracy of $\pm$\SI{0.60}{\centi\meter} in predicting the air column height for cylinders. Second, although the model is trained on cylinders, we show that the pitch estimation (and the model in general) is applicable beyond cylinders -- for example, it can be used to predict the shape of containers with convincing accuracy. Third, the model generalizes well to videos from other datasets and to in-the-wild YouTube videos.
In summary, our contributions are:
\begin{enumerate}
    \item We show in theory that the physical properties of the container-liquid system can be recovered from the fundamental frequency (pitch) of the sound of pouring.
    \item We train a pitch detection model using supervision from simulated data and visual data with a physics-inspired objective.
    \item We introduce a new clean and large dataset of videos of liquid pouring that can be used to study the estimation of physical properties.
    \item We demonstrate that the model is indeed capable of detecting pitch and hence estimates physical properties only from the sound of pouring. We show that the model generalizes to different container shapes, to other datasets, and to in-the-wild YouTube videos.
\end{enumerate}

\section{Related Work}

\paragraph{Psychoacoustics of pouring.}

Psychoacousticians have demonstrated a remarkable human ability to infer physical properties such as material~\cite{klatzky2000perception,traer2018intuitive}, shape~\cite{kunkler2000hearing}, and size~\cite{carello1998perception,rocchesso2003size,zwicker2013psychoacoustics} from sound alone. Furthermore, the temporal evolution of sound also provides an anticipatory sense in humans that helps to estimate dynamic properties such as distance~\cite{mershon1979absolute}, velocity~\cite{russell2023identifying}, and time-to-contact~\cite{gordon2013spectral}. Interestingly, liquid pouring presents a unique case in which humans can infer both static properties (e.g., container size, material, liquid temperature, etc.) and dynamic properties (e.g., liquid level, time-to-fill) from sound alone~\cite{Cabe2000HumanST,perfecto2019volume,velasco2013sound}.
We draw inspiration from this line of work and train audio models to infer physical properties from the sound of pouring.

\paragraph{Pouring in the literature.}
Pouring occurs surprisingly often in the literature.
In the machine learning community, pouring has been studied by roboticists to learn how to pour~\cite{pouring1_7989307,pouring2_Pan_Park_Manocha_2016,pouring3_HUANG2021103692,pouring4_huang2017learning,pouring5_8967802,pourin6_8967911,pouring7_9981195,pouring8_8593654, pouring9_9811898}. In computer vision, there has been much work on visually perceiving liquids either in a static setting~\cite{liquid_perception2_7759326,liquid_perception7_Mottaghi2017SeeTG,liquid_perception6_eppel2021predicting}, or during pouring~\cite{liquid_perception1_Narasimhan2022SelfsupervisedTL,liquid_perception3_schenck2016detection,liquid_perception4_Schenck2016TowardsLT,liquid_perception5_Lin_2023_ICCV,liquid_perception8_Schenck2017PerceivingAR}.
For example, \cite{liquid_perception2_7759326, liquid_perception7_Mottaghi2017SeeTG, liquid_perception1_Narasimhan2022SelfsupervisedTL} detect the amount of liquid in a container, \cite{liquid_perception6_eppel2021predicting} detect the container shape and material, and \cite{liquid_perception3_schenck2016detection, liquid_perception5_Lin_2023_ICCV} track the stream of pouring liquid.
Likewise, there has been work on estimating the dynamic latent states (e.g., height or mass of liquid at a given time) from multi-modal (vision, audition, haptics, etc.) inputs~\cite{Wilson2019AnalyzingLP,Liu2020VA2MassTT,Liang_2019,Liang_2020,zheng2021pouring,wu2018liquid}. The majority of these works~\cite{Liang_2019,Liang_2020,zheng2021pouring,wu2018liquid} use additional sensory data (e.g., force, torque, hand trajectory, inertial measurements) in combination with vision or audio or both. Such measurements require sophisticated recording equipment. In contrast, our aim is to be able to predict physical properties from sound alone,
and to achieve this without using bespoke equipment, but instead from regular smartphone recordings of liquid pouring.
Closest to our work is that of~\citet{Wilson2019AnalyzingLP}
where an audio-visual CNN is supervised to predict the mass of liquid poured at a given time, given instantaneous video and audio clips.
Methodologically, our work differs from \cite{Wilson2019AnalyzingLP} by incorporating the underlying physics directly in the learning process.
By design, our method can estimate several physical properties and not only the liquid mass without supervision.
We also evaluate our model by linear probing of the co-supervised features on the dataset of \cite{Wilson2019AnalyzingLP} and report superior performance.

\paragraph{Liquid and pouring datasets.} Most robotic pouring work uses private datasets recorded with a robotic arm in a lab setting, or simulated data to train robots. Although ~\cite{sermanet2018time, liang2019making,liang2020robust} records third-person pouring videos, these datasets are either not openly available, very small in scale, or have missing audio recordings. Pouring does feature in popular large-scale audio-visual datasets such as VGGSound~\cite{vggsound}, AudioSet~\cite{audioset} and EPIC-Kitchens~\cite{Damen2022RESCALING}. But these are very limited in number and too visually noisy (large viewpoint changes, low visibility, occlusions) for a controlled study.
Likewise, there are lots of pouring videos on YouTube (Shorts in particular), but these too are visually noisy. Closest to our requirements is the dataset collected by \citet{Wilson2019AnalyzingLP} of 500+ videos, but only
276 of them have liquid pouring
across only 4 containers.
So, we resort to recording our own dataset of 805 videos of pouring across 50 containers with a casual smartphone camera in a domestic setting. We will publicly release our dataset to stimulate further research. In addition, we also evaluate our models on the dataset proposed by \citet{Wilson2019AnalyzingLP}.

\paragraph{Audio-visual learning.}
The natural audio-visual correspondence in videos coupled with large-scale video datasets~\cite{audioset,vggsound} has led to an array of work on self-supervised representation learning for various downstream tasks
\cite{owens2018audio,gong2022contrastive,chen2021multimodal,georgescu2023audiovisual,shi2022learning,asano2020labelling}.
These approaches can be broadly categorized as contrastive~\cite{gong2022contrastive,ma2020active,akbari2021vatt}, generative~\cite{georgescu2023audiovisual,shi2022learning,huang2024mavil}, paired sample discriminative~\cite{arandjelovic2017look,owens2018audio,korbar2018cooperative,arandjelovic2018objects,morgado2021audio}, clustering~\cite{alwassel2020self,asano2020labelling,chen2021multimodal} and distillation-based~\cite{aytar2016soundnet,owens2016ambient}.
More recently, with the proliferation of
transformer-based language models,
audio representations have been learned together with text and video~\cite{akbari2021vatt,guzhov2022audioclip,zellers2022merlot,girdhar2023imagebind,gong2023listen}.
Beyond learning general representations, there has also been much work on special downstream tasks such as audio visual localization~\cite{arandjelovic2018objects,chen2021localizing,chen2022sound}, lip reading~\cite{afouras2018deep,Afouras18b}, sound/visual generation~\cite{du2023conditional,chung2024t,lee2022sound}. These approaches and tasks largely ignore the fine-grained time-dependent changes in sounds and rely on instantaneous and coarse correspondences.
In contrast, liquid pouring necessitates modeling of fine-grained characteristics over time (namely, pitch).

\section{The Physics of Liquid Pouring}

As an example case, consider a simple cylindrical vessel of radius $R$, height $H$ as shown in \cref{fig:resonance-sample} (a). At time $t$, suppose that the vessel is filled to a level such that the length of the air column is $l(t)$.
While water is poured, we hear a mix of pitch and odd harmonics that correlate with the length of the air column at a given time.
This is visible on the spectrogram in \cref{fig:resonance-sample} (b).
We term this resonance as \textit{axial} resonance.
The observed pitch is shown as the blue curve on the spectrogram in \cref{fig:resonance-sample} (d).
The coresponding wavelength that varies linearly w.r.t. $l(t)$ is shown in \cref{fig:resonance-sample} (c).
Here, we describe the physical equation that determines this curve and the physical properties derived from it. We also discuss another kind of resonance, termed \textit{radial} resonance, which is less prominent but co-occurs with axial resonance.

\subsection{Axial resonance}
\begin{figure}
    \centering
    \includegraphics[width=\textwidth]{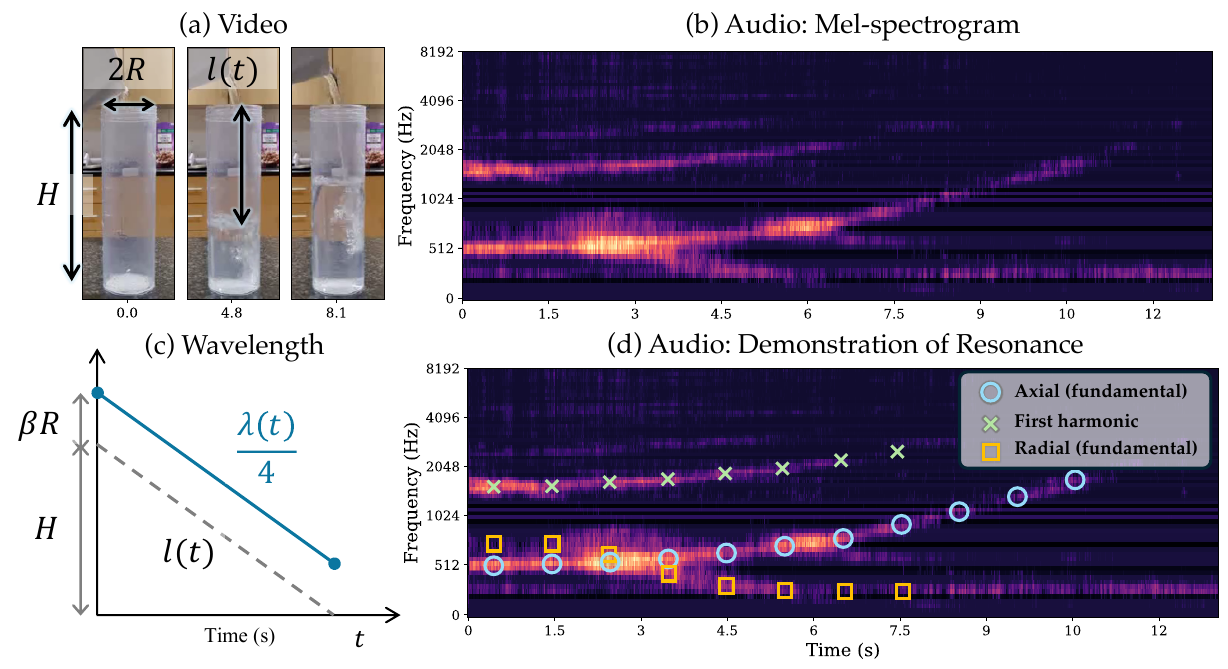}
    \captionsetup{font=normal,skip=1pt}
    \caption{
        \textbf{
            Demonstration of resonance in liquid pouring.
        } As liquid is poured in the container shown in (a) of height $H$ and radius $R$, a sound made up of an increasing pitch (fundamental frequency) and some (odd) harmonics is observed on the spectrogram shown in (b).
        Two kinds of resonance are observed: axial (fundamental shown as blue circles in (d), first harmonic as green crosses) and radial (fundamental shown as yellow squares). The wavelength (inverse of frequency) of the axial resonance (shown in (c)) is a function of the length of air column $l(t)$: $\lambda(t)/4 = l(t) + \beta R.$
        Interestingly, the high intensity blob around 3s is likely due to the mixture of pitch from both kinds of resonance.
    }
    \label{fig:resonance-sample}
\end{figure}

As the water fills up, it pushes out air in the air column creating a frequency pattern that resembles blowing air in an organ pipe closed at one end. This phenomenon has been studied by physicists for a long time~\cite{radial1-French1983InVV}. As the water level increases, the vacant space for air molecules to vibrate reduces and hence the frequency increases. This frequency can be described mathematically with an analogy to a string tied at one end. At time $t$, the fundamental frequency $f(t)$ is given by
\begin{equation}
    f(t) = \frac{c}{4} \frac{1}{l(t)},
\end{equation}
where $c$ is the speed of sound in air. This expression arises from a standing wave of wavelength $\lambda(t) = 4 l(t)$ where the amplitude is zero at the water surface and maximum at the top of the vessel.
Rayleigh~\cite{rayleigh1896theory} and others studied this and found an experimental end-correction that depends on the radius of the container:
\begin{equation}
    f(t) = \frac{c}{4} \frac{1}{\left( l(t) + \beta R \right)},
\end{equation}
where $\beta$ is an end-correction factor generally agreed to be $0.62$~\cite{Anderson1928EffectOF}.
The numerical value of $\beta$ has also been debated in the acoustic-physics community~\cite{Anderson1928EffectOF,Jones1941EndCO,pykett_organ_pipes} but we fix it to $0.62$.
A spectrogram of the sound of pouring in a sample container is shown in \cref{fig:resonance-sample} (b). The observed pitch $f(t)$ (blue circles) and first harmonic (green crosses) are marked on the spectrogram in \cref{fig:resonance-sample} (d).

To avoid working with an inverse relation, we look at this equation in terms of wavelength $\lambda(t)$,
\begin{equation}
    \lambda(t) = \frac{c}{f(t)} =  4\left(l(t) + \beta R\right).
    \label{eq:wavelength-length}
\end{equation}
Note that all these quantities are in metric units. The LHS is observable from the audio of liquid pouring while the RHS is observable from a video of liquid pouring (up to a scale factor).
Fascinatingly, this implies that the audio is effectively a \textit{metric ruler} for objects in the video.

\subsection{Radial resonance}
\label{subsec:radial}

In addition to axial resonance, another kind of resonance is observed when water is poured into a vessel. This is subtle and not directly visible or even clearly audible, but it does show up when carefully analyzing spectrograms. As water is poured, it generates vibrations on the surface of the container and the container vibrates radially. As the liquid level increases, the mass of the combined container-liquid system increases, which acts as an inertia that decreases the vibration frequency. Thus, in this case, as the liquid level increases, the frequency decreases. This has been studied in~\cite{radial1-French1983InVV,radial2-PhysRev.13.171} and is also related to the notion of musical glasses~\cite{Guignard2003TuningOM}. Formally, this frequency for cylindrical containers is given by~\cite{radial1-French1983InVV}:
\begin{equation}
    \label{eq:radial-frequency}
    f(t) = \frac{f_{0}}{\left[ 1 + \xi \left(1 - \frac{l(t)}{H}\right)^{3}  \right]^{1/2}},
\end{equation}
where $f_{0}$ depends on the container's dimensions and physical properties such as  density, thickness, and $\xi$ depends on the liquid density as well as container material density. In this case, the observed frequency curve is shown with yellow squares in \cref{fig:resonance-sample} (d).

\subsection{Recovering physical properties from pitch}
\label{subsec:theory}

Our main observation is that to determine
some of the key
physical properties, it suffices to estimate the \textit{fundamental wavelength} in {\bf axial} resonance from raw audio.
Recall that the basic physical relation driving this system is given by \cref{eq:wavelength-length}.
We categorize the physical properties in two sets. (i) \textit{Static properties}: these are
inherent to the container-liquid system (e.g., container size) and do not vary over time. (ii) \textit{Dynamic
    properties}: these are a function of time (e.g., length of the air column, flow rate, time to fill).
We start with the derivation for the length of air column and then derive other properties from that.

\begin{enumerate}[label=(\roman*)]
    \item \textbf{Length of air column}: We want to estimate $l(t)$ given $\lambda(t)$ at a given time $t$. If we also know $l, \lambda$ at another point $t' \neq t$, then we get:
          \[
              l(t) = l(t') + \frac{1}{4}\left[\lambda(t) - \lambda(t') \right].
          \]
          Using the boundary condition $l(T) = 0$, where $T$ is the total pouring duration, we get:
          \begin{equation}
              \label{eq:length}
              l(t) = \frac{1}{4}\left[\lambda(t) - \lambda(T)\right]
          \end{equation}

    \item \textbf{Container size}: Container height and radius are directly obtained from the boundary conditions:
          \begin{equation}
              \label{eq:container-dimensions}
              H = l(0) = \frac{\lambda(0) - \lambda(T)}{4} \ \ \text{and} \ \ R = \frac{\lambda(T)}{4\beta}
          \end{equation}
    \item \textbf{Volume flow rate}: Likewise, we can derive the other properties. For the volume flow rate $Q(t)$, suppose the volume at time $t$ is $V(t)$. Then,
          \begin{equation}
              \label{eq:flowrate}
              Q(t) = \dv[]{V}{t} = \pi R^{2} \dv[]{(H - l(t))}{t} = -\pi R^{2}\dv[]{l}{t} = -\frac{1}{4}\pi R^{2}\dv[]{\lambda}{t},
          \end{equation}
          where the derivative $\dv[]{\lambda}{t}$ can be approximated using the estimated $\lambda(t)$.
    \item \textbf{Time to fill}:
          For time to fill, we assume a constant flow rate (since otherwise, one could pause pouring midway leading to ill-defined time to fill). Also, we do not know the true duration $T$ and are only given a partial audio, i.e. cut upto time $t$. Here, following \citet{Cabe2000HumanST}, we make an additional assumption that the end-correction term $\beta R$ is small at the start of pouring ($\beta R \ll H$). Thus, in a short interval at the start of pouring $t' \in (0, \delta)$, ignoring the end correction, we can approximate
          \begin{equation}
              \label{eq:ttf}
              \tau(t') = -\left[ \frac{l(t')}{\dv[]{l}{t}} \right] = -\left[ \frac{\lambda(t') - \beta R}{\dv[]{\lambda}{t}} \right] \approx -\frac{\lambda(t')}{\dv[]{\lambda}{t}}, \forall t' \in (0, \delta).
          \end{equation}
          Then, we can use the property of $\tau$ to get time to fill at any given time $t$:
          \[
              \tau(t) = T - t = (T - t') + t' - t = \tau(t') - (t - t'),
          \]
          for some $t' \in (0, \delta), \delta \ll t$. Note that this needs reliable estimates of $\lambda$ and its derivative at the beginning of the audio.
\end{enumerate}

Hence, we have shown that to estimate the desired physical properties, it suffices to determine the fundamental wavelength $\lambda(t), \forall t.$ Note that we require precise estimation of $\lambda(t), \forall t \in [0, T]$ and particularly at the start ($t=0$) and end ($t=T$).

\section{Audio Network and Training}
Our objective is to predict physical properties (\eg, length of air column, size of the container) from the sound of pouring.
Our approach is formulated as a two-stage process: (i) detecting the pitch from the raw audio signal, and (ii) recovering these physical properties from the pitch.
To detect pitch, we propose a network based on \texttt{wav2vec2} (\cref{subsec:audio-network-arch}). We pre-train it on simulated sounds of liquid pouring (\cref{subsec:pretraining-with-synth}). Then, on real data, we use the visual stream to co-supervise the audio network (\cref{subsec:ft-cosupervision}). Given a strong pitch detector, we use \cref{eq:container-dimensions,eq:length,eq:flowrate,eq:ttf} to obtain the desired physical properties from the pitch.

\subsection{Audio network architecture}
\label{subsec:audio-network-arch}

\begin{figure}[h]
    \centering
    \includegraphics[width=\textwidth]{
        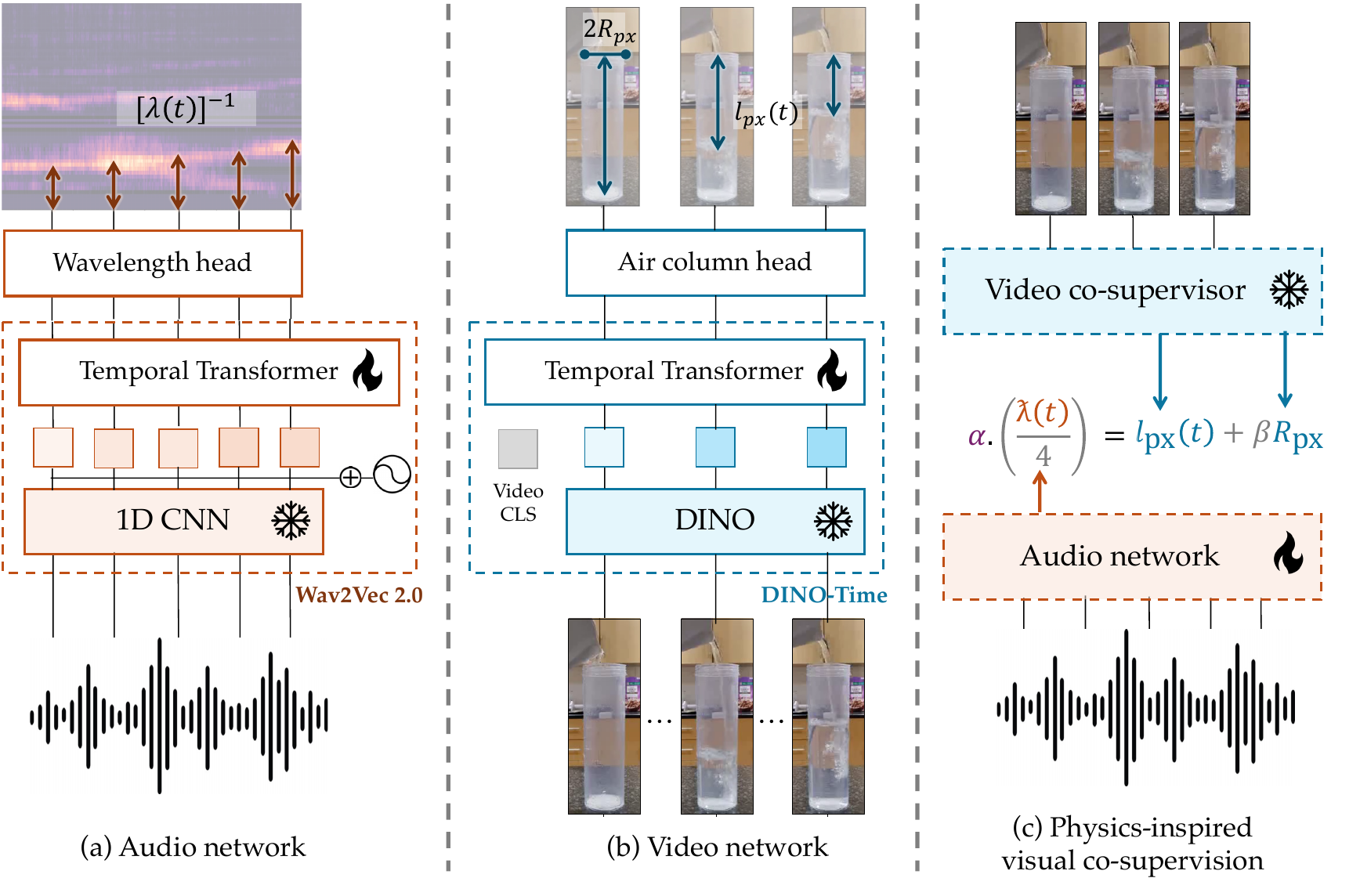
    }
    \captionsetup{font=normal,skip=1pt}
    \caption{
        \textbf{Model architecture and training.}
        (a) The \colorbox{modorange!30}{audio network} is based
        on a \texttt{wav2vec2} repurposed for pitch detection.
        (b) The \colorbox{modblue!30}{video network} is based on
        DINO repurposed to operate on image sequences to detect
        length of air column and container radius (up to a scale
        factor). (c) The audio network is pre-trained on
        synthetic samples and then fine-tuned on real samples
        using physics-inspired co-supervision from the video.
    }
    \label{fig:stage1-arch}
\end{figure}

The network takes in raw audio samples and outputs wavelength (pitch) estimates at each time step.
The architecture is based on \texttt{wav2vec2}~\cite{baevski2020wav2vec} adapted for pitch detection on pouring sounds. 
The architecture diagram is shown in \cref{fig:stage1-arch} (a).
The network takes in raw audio waveform and outputs a distribution over the set of wavelength bins at each time step.
Note that we predict the fundamental wavelength as opposed to fundamental frequency because the wavelength varies linearly with the length of the air column and we want to bake in this linearity in the learned features.
The input waveform is resampled at a rate of 16k Hz.
First, the waveform is tokenized using a 1D CNN encoder which takes in windows of 400 samples (25ms) of audio with a hop length of 320 samples (20ms).
In addition to the original model design, we add sinusoidal position embeddings to the tokens to enhance temporal information.
These are then passed through a Transformer network with 12 blocks (model dimension 768, 8 attention heads).
This is followed by a prediction head, a linear regressor that maps from $\mathbb{R}^{768} \rightarrow \mathbb{R}^{K}$ where $K$ is the number of wavelength bins. The output is converted into a distribution over wavelength using Softmax activation.
More details on the architecture are provided in \cref{appendix-subsec:audio-network}.

\paragraph{Training the network.} We want to train the described network to detect pitch. Since it is difficult to obtain pitch annotations on real samples, we first pre-train the network on synthetic samples with perfect ground truth (\cref{subsec:pretraining-with-synth}). Then, we fine-tune on a small amount of real data with video as the source of co-supervision (\cref{subsec:ft-cosupervision}).

\subsection{Pre-training with synthetic data}
\label{subsec:pretraining-with-synth}

First, we describe how we generate simulated pouring sounds and then, we describe the pre-training details.

\begin{figure}[t]
    \centering
    \includegraphics[width=\linewidth]{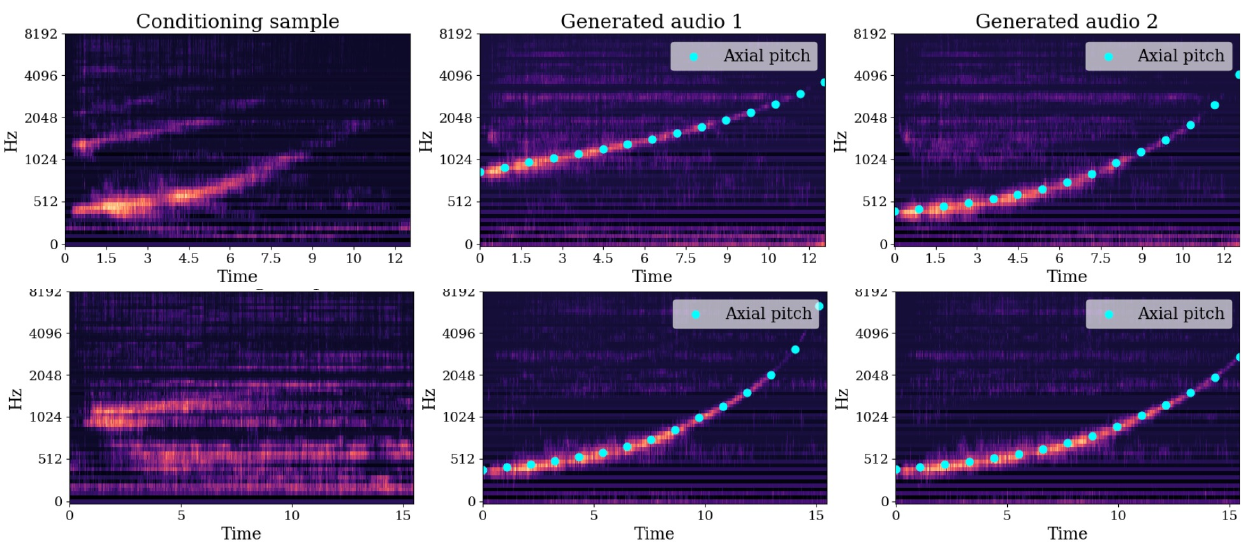}
    \captionsetup{skip=2mm}
    \caption{
        \textbf{Samples of simulated pouring sounds.}
        Our simulator takes in (i) a real sample from the train set as condition, (ii) a random pitch profile and generates synthetic waveform that resembles sound of pouring liquid in a cylindrical container. More samples are shown in \cref{appendix-subsec:audio-network}.
    }
    \label{fig:synthetic-samples-main}
\end{figure}

\paragraph{Synthetic data generation.}
We train a generative model based on Differentiable Digital Signal Processing (DDSP)~\cite{engel2020ddsp} to simulate the sounds of liquid pouring.
The architecture is a supervised autoencoder model with the latent space decomposed into pitch, loudness, and a residual vector.
The encoder represents the raw audio waveform in three parts: (i) pitch over time, (ii) loudness over time, and (iii) residual vector.
The pitch is extracted using CREPE~\cite{kim2018crepe}, a standard pitch detector popularly used in music.
Loudness is extracted as standard RMS energy.
The residual is learned with a GRU network operating on Mel Frequency Cepstrum Coefficients (MFCC) features.
Intuitively, the residual captures background noise and room reverberation characteristics.
The decoder is composed of synthesizers based on classical signal processing techniques.
It takes in pitch, loudness, and the residual and generates a realistic waveform.
The network is trained using a multi-scale spectrogram reconstruction loss.
To generate a sample, we first randomly pick a conditioning sample from the train set and extract loudness and residual vector from it.
We can pass an arbitrary pitch profile with the chosen loudness and residual.
To simute a pitch profile, we sample dimensions of an arbitrary cylindrical container. Formally, we sample radius $R$, height $H$, and then compute length of air column as a linear curve:
\begin{equation}
    l(t) = \left(-\frac{H}{T}\right)t + H,
\end{equation}
where $T$ is the duration of the conditioning sample.
Then, we plug this in \cref{eq:wavelength-length} to compute the wavelength $\lambda(t)$ which is inverted to obtain the pitch.
These three (pitch, loudness, residual) are then fed to the trained decoder which generates a realistic waveform with the desired pitch profile.
For a single conditioning sample, the loudness and residual are fixed while we can vary $(R, H)$ to vary the pitch and cover a vast diversity of pitch profiles.
Some examples of generated samples are shown in \cref{fig:synthetic-samples-main}.
We sample $H \sim U[5, 25]$ cm, $R \sim U[1, 5]$ cm and generate $10{,}000$ samples. This is randomly divided into a train and validation set.

\paragraph{Implementation details for pre-training.} We train only the penultimate 8 layers of the Transformer and the prediction head keeping the rest of the network frozen.
The number of wavelength bins is chosen as $K{=}64$. The wavelength range is chosen to be $[0, L], L {=} 100$.
Each bin represents a length of $L/K {=} 100/64 {=}1.56$ cm.
Following \cite{kim2018crepe}, to soften the penalty for near-correct predictions,
the target is Gaussian-blurred in wavelength such that the energy surrounding a ground truth wavelength decays with a standard deviation of 1.25 bins (or 1.95 cm). The network is trained with a KL divergence penalty between the predicted and true distributions. It is trained for 100 epochs with a batch size of 32 using Adam optimizer~\cite{kingma2014adam} with a constant learning rate of $1e^{-4}$.

\subsection{Fine-tuning by visual co-supervision}
\label{subsec:ft-cosupervision}

The pre-trained audio network needs to overcome the sim2real gap to detect pitch accurately on real samples. Since it is difficult to obtain ground-truth pitch annotations on real samples, we make use of the video stream as a source of weak supervision for fine-tuning. From the RHS in \cref{eq:wavelength-length}, the video can supply measurements for the length of the air column and the container radius (up to a scale), which we use to supervise the audio network.

\paragraph{Video pre-training with pseudo labels.} To use video as a teacher for co-supervision, we pre-train a video network to detect the length of the air column (or equivalently, the level of liquid) and the average radius of the container.
The architecture is shown in \cref{fig:stage1-arch} (b).
We use a frozen DINO~\cite{caron2021emerging} encoder per frame and attach a Transformer to model the temporal dependencies. This is followed by a prediction head for the length of the air column $l(t)$ (relative to the image size). The (psuedo) labels to train this network are obtained using temporal difference between adjacent frames and classical image processing techniques (Derivative of Gaussian on temporal difference heatmaps) to obtain clean ground truths.
More details are provided in \cref{appendix-subsec:visual-network}.
The radius of the container is obtained by segmenting the container from the first frame of the video using SAM~\cite{kirillov2023segment}. Note that both these measurements are in pixel scale.

\paragraph{Scale-aware video co-supervision.} The wavelength predictions $\lambda(t)$ from the audio network are in metric units while the length predictions $l_{px}(t), R_{px}$ from the video network are in pixels. To enable video as a weak supervisor, we need to account for a scale factor, $\alpha$.
\begin{equation}
    \alpha. \left(\frac{\lambda(t)}{4}\right) = l_{px}(t) + \beta R_{px}
    \label{eq:cosupervision}
\end{equation}
This factor encapsulates the depth and camera intrinsics and is unique per video. Assuming a usual perspective camera, we can obtain a precise definition of the scale factor:
\begin{equation}
    \alpha := \frac{1}{Z}\frac{f}{s},
    \label{eq:alpha}
\end{equation}
where $Z$ is the depth, $f$ is the focal length and $s$ is length per pixel. Note that $\alpha$ is inversely proportional to depth.
We can pre-compute $\alpha$ for each video by simply computing the ratio of wavelength from audio to the pixel lengths from video. However, since the audio estimates are generally off where the signal is low (e.g., towards the end of the pouring sequence), this leads to very poor estimates of $\alpha$. To account for this, we weigh the ratios over time with the RMS energy in the audio over time.
This leads to robust scale estimates. Some example containers and scale estimates are shown in \cref{appendix-subsec:visual-network}.
For example, a large container
is kept further away from the camera and thus has a smaller $\alpha$.
Next, we fix the scale factors and the video network and fine-tune the audio network with the MSE loss to improve predictions of $\lambda(t)$ using supervision based on \cref{eq:cosupervision}.

\paragraph{Implementation details for fine-tuning.} We convert the audio network outputs from distributions to scalar wavelengths. The video outputs are already scalars. We use the standard MSE loss (\cref{eq:cosupervision}) and fine-tune the network for 5 epochs with the Adam optimizer~\cite{kingma2014adam} and a constant learning rate of $1e^{-6}$. At inference, given a sound of pouring, our model predicts a wavelength distribution which is converted to wavelength scalars and subsequently to frequency scalars. These frequencies can then be overlaid on spectrograms to verify correctness qualitatively.

\section{The \textit{Sound of Water 50} Dataset}
\begin{figure}[t]
    \centering
    \includegraphics[width=\linewidth]{
        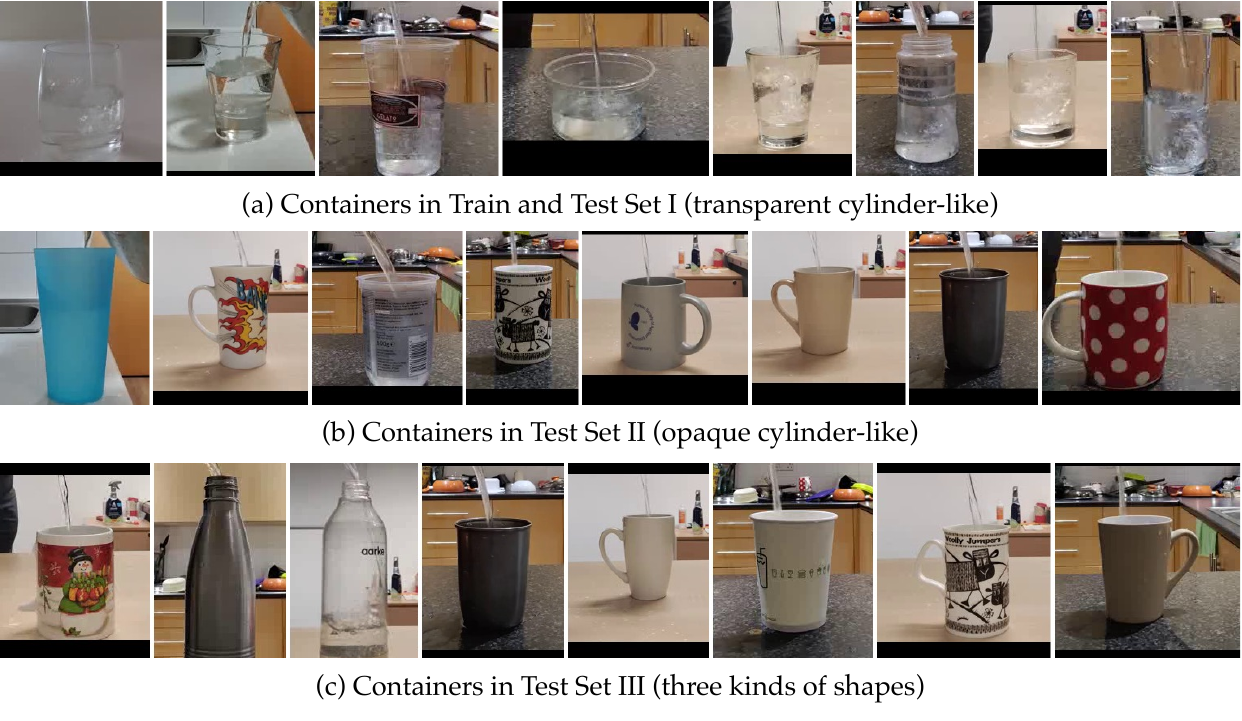
    }
    \captionsetup{skip=0mm}
    \caption{
        \textbf{Examples of containers used in the \textit{Sound of Water 50} dataset.}
        The dataset contains videos of pouring liquids in
        containers with diverse shapes, materials, opacity
        and background environments. The train set has videos
        of pouring in transparent cylinder-like containers.
        Test set I shares the same set of containers but is
        distinct in terms of the videos. Test set II and II
        have videos of pouring in entirely unseen containers.
    }
    \label{fig:container-samples}
\end{figure}

\begin{figure}
    \centering
    \includegraphics[width=\linewidth]{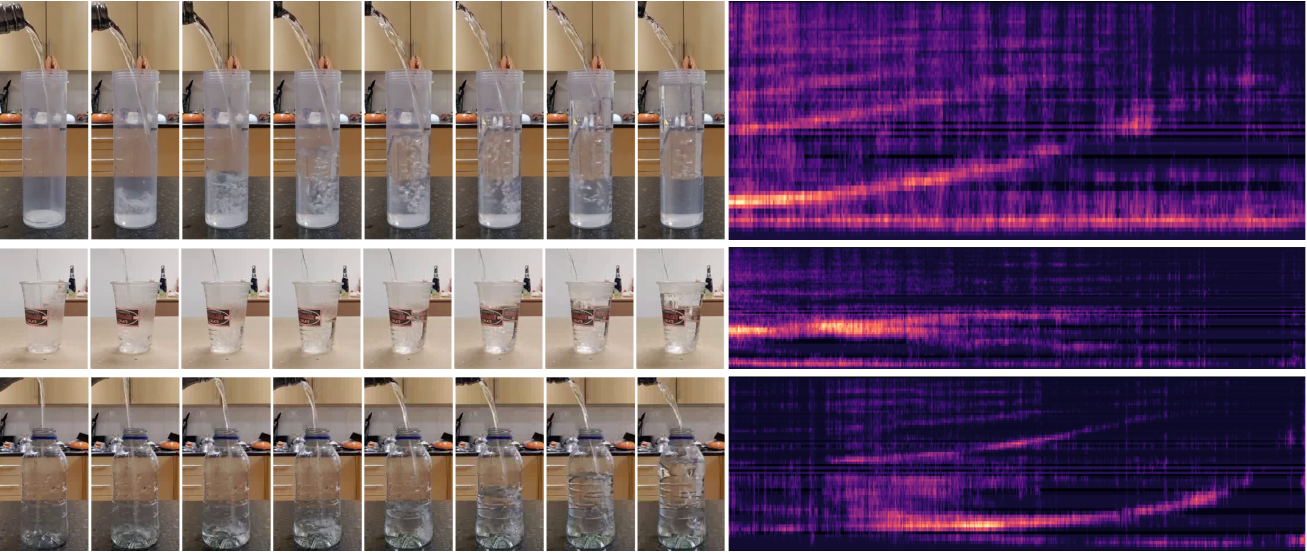}
    \caption{\textbf{Sample video sequences from the \textit{Sound of Water 50} dataset.} (Left) Sample pouring sequences in each of the three container shapes: cylindrical, semiconical and bottleneck.
    (Right) The corresponding spectrograms of the pouring sounds.
    }
    \label{fig:sequences-sample}
\end{figure}

Our dataset consists of videos showing a human hand pouring liquid in a container with a fixed camera facing the container.
The videos are recorded by the authors with a smartphone camera in a domestic setting.
Across videos, we randomly vary the flow rate but keep it approximately constant within a single video. In total, we collect 805 videos across 50 containers (4 shapes, 5 materials) and 2 liquids (hot and normal water).
The shapes are cylindrical, semiconical, bottleneck, and hemispherical.
The materials include glass, plastics, ceramics, steel, and cardboard.
Some example containers are shown in
\cref{fig:container-samples}. Some example video sequences are shown in \cref{fig:sequences-sample}.

\paragraph{Splits.}
Since we only train with cylinder-shaped containers, we carefully create splits with a single training set and multiple test sets. Details of all four splits are provided in \cref{tab:data-splits}.
In short, the train set has videos of pouring in transparent cylinder-like containers.
Test set I has a subset of containers of those in test (seen containers) but is distinct in terms of video sequences.
Test set II and III have completely unseen containers.
We use the different test sets according to the requirements of a given task, e.g., we use \textit{Test Set III} to probe shape classification.
Some examples of containers in each split are shown in \cref{fig:container-samples}.


\begin{table}[ht]
    \centering
    \resizebox{\columnwidth}{!}{%
        \begin{tabular}{lcccccccl}
            \toprule
            \textbf{Split}    & \multicolumn{2}{c}{\textbf{Opacity}} & \multicolumn{3}{c}{\textbf{Shapes}} & \textbf{\# containers}   & \textbf{\# videos}        & \multicolumn{1}{c}{\textbf{Description}}                                                                                      \\
            \cmidrule{2-6}
                              & {\footnotesize Transparent}          & {\footnotesize Opaque}              & {\footnotesize Cylinder} & {\footnotesize Semi-cone} & {\footnotesize Bottle}                   & \multicolumn{1}{l}{} & \multicolumn{1}{l}{} &                                      \\
            \midrule
            \textit{Train}    & \cmark                               & \xmark                              & \cmark                   & \cmark                    & \xmark                                   & 18                   & 195                  & Transparent cylinder-like containers \\
            \textit{Test I}   & \cmark                               & \xmark                              & \cmark                   & \cmark                    & \xmark                                   & 13                   & 54                   & Test set with seen containers        \\
            \textit{Test II}  & \xmark                               & \cmark                              & \cmark                   & \cmark                    & \xmark                                   & 19                   & 327                  & Test set with unseen containers      \\
            \textit{Test III} & \cmark                               & \cmark                              & \cmark                   & \cmark                    & \cmark                                   & 25                   & 434                  & Shape clf.\ with unseen containers   \\
            \bottomrule
        \end{tabular}%
    }
    \caption{
        \textbf{Splits in \textit{Sound of Water 50} dataset.}
        We create four splits in our dataset. The train set
        contains videos of pouring in transparent cylinder-like
        containers. Test set I contains containers that are a
        subset of those in train set but has distinct video sequences.
        Test set II has completely unseen cylinder-like containers.
        Test set III also has unseen containers including bottleneck
        shaped containers. Test III is used for shape classification
        and overlaps only with Test II in terms of containers.
        This adds up to $18+25 = 43$ containers, and $195+54+434 = 683$
        videos. The remaining 122 videos (out of a total of 805) are of
        hemispherical/freeform containers only used for qualitative analysis.
        Some example containers in each of these sets are shown
        in \cref{fig:container-samples}.
    }
    \label{tab:data-splits}
\end{table}
\begin{figure}[ht]
    \centering
    \includegraphics[width=\linewidth]{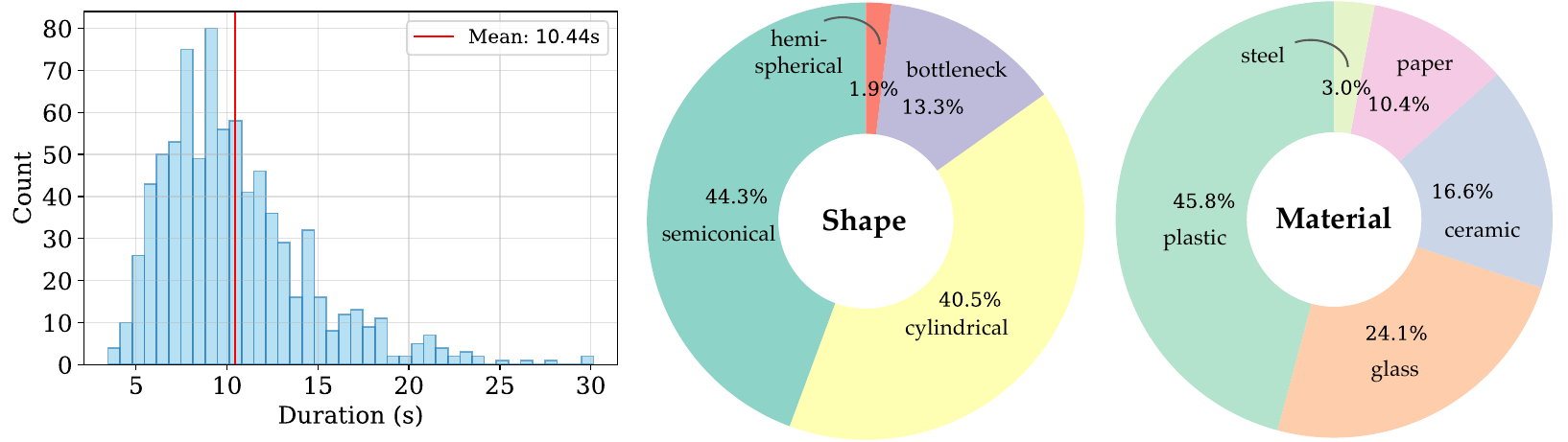}
    \caption{\textbf{\textit{Sound of Water 50} dataset  statistics.} We show some basic numbers from the dataset. (Left) shows distribution of video duration, (center) shows distribution over container shapes and (right) shows distribution over container material.}
    \label{fig:data-stats}
\end{figure}

\paragraph{Statistics.}
Basic statistics from our dataset are shown in \cref{fig:data-stats}.
Since we use a variety of container sizes, the video duration ranges from \SI{3.5}{\second} to \SI{30.1}{\second} with a mean of \SI{10.4}{\second}.
The average height and (base) radius of a container are \SI{11.5}{\centi\meter} and \SI{3.1}{\centi\meter}, respectively.
The containers are sourced from four shapes (cylindrical, semi-conical, bottleneck, hemispherical) and five materials (glass, plastic, steel, paper, ceramic) distributed as shown in \cref{fig:data-stats}.

\section{Experiments}

In this section, we present various experiments to demonstrate physical understanding from sounds of liquid pouring.
We start with a brief description of our dataset.
Then, we describe the results on estimating pitch (\cref{tab:pitch-detection-results}) and physical properties from pitch (\cref{tab:physical-properties}). Then, we present results on classifying container shapes and estimating liquid weight directly from the learned representations.

\subsection{Estimating basic physical properties from pouring sound}
\label{subsec:physical-properties}

As theoretically shown in \cref{subsec:theory}, we now experimentally show estimation of physical properties from pouring sound.

\paragraph{Evaluation on air column length.}
We evaluate our pitch detection network to estimate the air column length.
We compare against simpler baselines for pitch detection. We try classical pitch detectors like Yin~\cite{de2002yin}, CNN-based detectors like CREPE~\cite{kim2018crepe} and recent self-supervised detectors like PESTO~\cite{riou2023pesto}. We also try a baseline with per-frame argmax on the spectrogram. Note that other baselines estimate pitch directly from raw audio samples while the argmax method operates on a spectrogram.
Given pitch (fundamental wavelengths) at certain time points, we fit a line to estimate $\lambda(t)$ on the collection of obtained points using RANSAC to avoid outliers.

To compute length of air column from wavelengths, we rely on \cref{eq:length}.
We compare our models with the baselines in estimating $l(t)$ and report the mean absolute error averaged over all time points.
The results are reported in \cref{tab:pitch-detection-results}.
Our models outperform all the baselines. Moreover, the co-supervised model achieves an error of 0.60 cm as compared to the audio-only variant's 0.78 cm on Test set I. The results are similar on the more challenging Test set II.

\paragraph{Evaluation on other physical properties.}
Having shown the superiority of our pitch detection model, we now check how much visual co-supervision helps in estimating other physical properties in \cref{tab:physical-properties}.

\begin{figure}[ht]
    \centering
    \includegraphics[width=\textwidth]{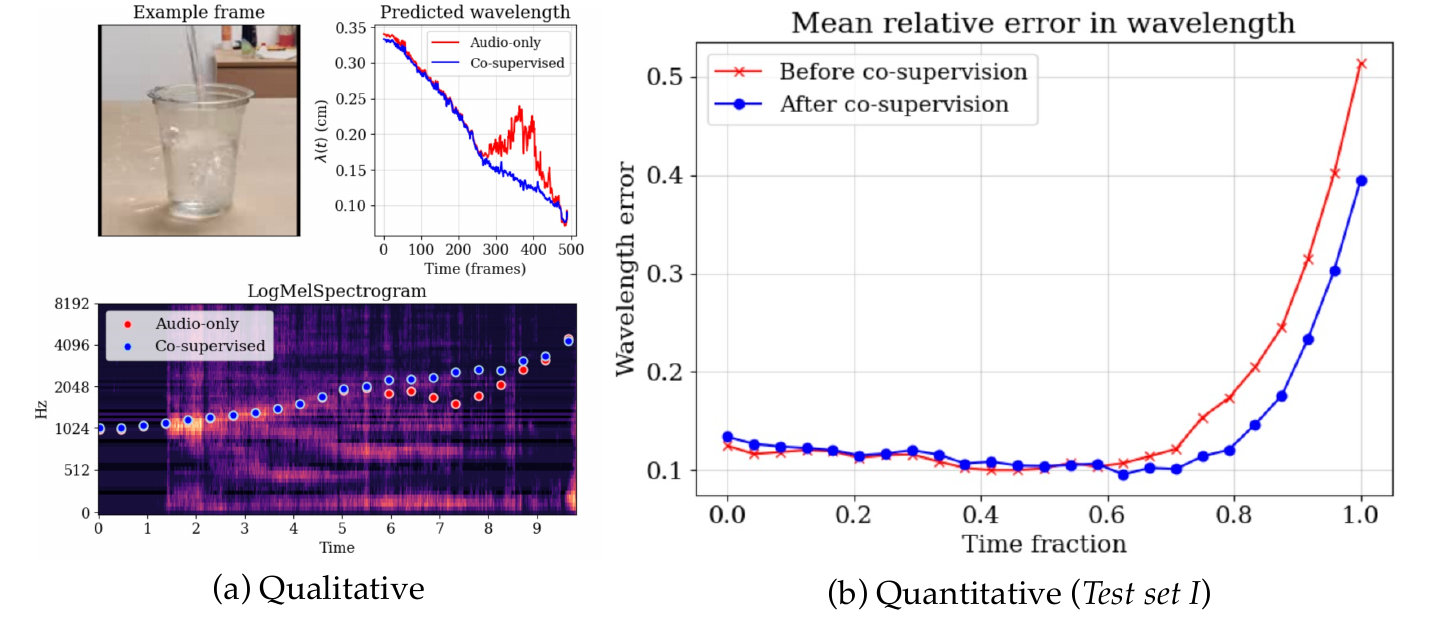}
    \captionsetup{font=normal,skip=1pt}
    \caption{\textbf{Where does co-supervision help the most?} We find that co-supervision is most beneficial towards the end of the audio where the signal is generally low. This is shown qualitatively on a sample in (a) and quantitatively on Test set I in (b). This helps in more precise radius estimation since it depends on $\lambda(T)$. Moreover, qualitatively, and quantitatively in estimating flow rate, it seems to help with more reasonable slopes of $\lambda$.}
    \label{fig:cosupervised-vs-audioonly}
\end{figure}

\begin{itemize}[leftmargin=5mm]
    \item \textbf{Container dimensions.} We obtain the container radius $R$ and height $H$ using \cref{eq:container-dimensions}. The ground truths are obtained by manually measuring the said containers. Our best model achieves an MAE of 2.27 cm in height and 1.39 cm in radius on Test set I and 2.77 cm / 1.88 cm on Test set II. Note that radius measurement depends on reliable wavelength estimates towards the end of the audio which is significantly better in case of a co-supervised model than the audio-only variant. This is further shown qualitatively and quantitatively in \cref{fig:cosupervised-vs-audioonly}.
    \item \textbf{Volume flow rate.} Following \cref{eq:flowrate}, we evaluate volume flow rate prediction. This is dependent on $R$ and derivative of $\lambda(t)$. The ground truth is obtained using the ratio of the volume of container and the time it took to fill it completely. The co-supervised model achieves 22.5 ml/s compared to 25.2 ml/s of the audio-only model on Test set I. Similar improvement is observed on Test set II.
    \item \textbf{Time to fill.} Following \cref{eq:ttf}, we evaluate on time to fill prediction. We are given a partial audio, i.e. cut at 25\% or 50\% or 75\% of its original duration and the task is to predict time (in seconds) that it will require to fill the container. Since we vary the flow rates randomly, it prevents the obvious shortcut of simply extrapolating based on the given time. As shown in \cref{tab:physical-properties}, the co-supervised model is comparable to the audio-only when provided with only 25\% of the input but outperforms the latter given 50\% or 75\% of the input on Test set I. On Test set II, the co-supervised model outperforms at all levels of the input.
\end{itemize}

\begin{table}[ht]
    \centering
    \resizebox{0.72\columnwidth}{!}{%
        \begin{tabular}{lcc}
            \toprule
            \textbf{Method}            & \begin{tabular}[c]{@{}c@{}}\textbf{Test set I} \\ seen containers $\downarrow$ \end{tabular} & \begin{tabular}[c]{@{}c@{}}\textbf{Test set II}\\ unseen containers $\downarrow$\end{tabular} \\ \midrule
            \multicolumn{3}{l}{\cellcolor[HTML]{f2f2f2}\textit{\textbf{Baselines}}}                                                                                                                                                   \\[1mm]
            Yin~\cite{de2002yin}       & 30.80                                                                                        & 27.30                                                                                         \\
            PESTO~\cite{riou2023pesto} & 11.70                                                                                        & 10.60                                                                                         \\
            CREPE~\cite{kim2018crepe}  & 7.61                                                                                         & 9.40                                                                                          \\
            argmax on spectrogram      & 4.60                                                                                         & 5.11                                                                                          \\
            \multicolumn{3}{l}{\cellcolor[HTML]{f2f2f2}\textit{\textbf{Ours}}}                                                                                                                                                        \\[1mm]
            Audio-only                 & 0.78                                                                                         & 0.82                                                                                          \\
            Co-supervised              & \textbf{0.60}                                                                                & \textbf{0.71}                                                                                 \\
            \bottomrule
        \end{tabular}%
    }
    \caption{
        \textbf{Comparison with baselines in estimating length of air column.}
        Mean absolute error (in cms) in estimating the length of air
        column $l(t)$ on the two test sets. Our models comfortably
        beat all the pitch detection baselines. Generally,
        performance on Test set II is lower as it consists of
        containers not seen during training.
    }
    \label{tab:pitch-detection-results}
\end{table}

\begin{table}[ht]
    \centering
    \resizebox{\columnwidth}{!}{%
        \begin{tabular}{lrrrrrr}
            \toprule
            \textbf{Property}              & \textbf{Units} & \textbf{Notation}       & \multicolumn{2}{c}{\textbf{Test set I}}       & \multicolumn{2}{c}{\textbf{Test set II}}                                                                                                              \\ \cmidrule{4-6}
            \textbf{}                      & \textbf{}      & \textbf{}               & \textit{\footnotesize Synthetic} $\downarrow$ & \textit{\footnotesize Co-supervised} $\downarrow$ & \textit{\footnotesize Synthetic} $\downarrow$ & \textit{\footnotesize Co-supervised} $\downarrow$ \\
            \midrule
            \multicolumn{7}{l}{\cellcolor[HTML]{f2f2f2}\textit{\textbf{Static properties}}}                                                                                                                                                                                                   \\[1mm]
            Height                         & cm             & $H$                     & 2.23                                          & \cellcolor[HTML]{FFE8E6}2.27                      & 2.77                                          & \cellcolor[HTML]{FFE8E6}2.85                      \\
            Radius                         & cm             & $R$                     & 1.62                                          & \cellcolor[HTML]{E8FFD9}1.39                      & 2.24                                          & \cellcolor[HTML]{E8FFD9}1.88                      \\
            \midrule
            \multicolumn{7}{l}{\cellcolor[HTML]{f2f2f2} \textit{\textbf{Dynamic properties}}}                                                                                                                                                                                                 \\[1mm]
            Flow rate                      & ml/s           & $Q(t)$                  & 25.20                                         & \cellcolor[HTML]{E8FFD9}22.50                     & 45.70                                         & \cellcolor[HTML]{E8FFD9}40.41                     \\
                                           & s              & $\tau_{\frac{1}{4}}(t)$ & 3.96                                          & \cellcolor[HTML]{FFE8E6}4.16                      & 4.39                                          & \cellcolor[HTML]{E8FFD9}4.10                      \\
                                           & s              & $\tau_{\frac{1}{2}}(t)$ & 1.62                                          & \cellcolor[HTML]{E8FFD9}1.49                      & 3.44                                          & \cellcolor[HTML]{E8FFD9}2.99                      \\
            \multirow{-3}{*}{Time to fill} & s              & $\tau_{\frac{3}{4}}(t)$ & 1.53                                          & \cellcolor[HTML]{E8FFD9}1.07                      & 2.66                                          & \cellcolor[HTML]{E8FFD9}2.21                      \\
            \bottomrule
        \end{tabular}%
    }
    \caption{
        \textbf{Co-supervision improves physical property estimation.}
        Mean absolute error in estimating various physical properties.
        Our visually co-supervised model generally improves over the
        synthetic-trained model in estimating physical properties from
        pitch. We observe noticeable improvements in estimating radius
        and flow rate. This suggests co-supervision particularly
        improves estimation of pitch towards the end of the audio
        as well as the slope of pitch generally.
    }
    \vspace{-0.5em}
    \label{tab:physical-properties}
\end{table}

\subsection{Recognizing liquid container shape from its pouring sound}

Differently shaped containers exhibit different structures of the resonance curves. Sound of pouring in cylindrical containers show a pitch that varies as $1/l(t)$ where $l(t)$ is the length of air column~\cite{Cabe2000HumanST}. In contrast, in bottles, pitch varies as $1/\sqrt{l(t)}$~\cite{webster2010use}. Though we train for pitch detection only on cylinder-like containers, here we evaluate whether the learned representations encode container shape. Formally, given the sequence of features of the sound of pouring, the task is to classify whether the shape of the container is cylindrical, semiconical, or bottleneck.

Given the sound of pouring, we first compute features from the co-supervised audio encoder. These are outputs, say $\{\mathbf{z}_{i}\}_{i=1}^{N}$, from the last Transformer block of \texttt{wav2vec2} before going into the prediction head in \cref{fig:stage1-arch}.
Then, we concatenate the following vectors to create a single summary vector per sample: the mean of the sequence and each of the vectors at 25\%, 50\% and 75\% of the sequence length.
\begin{equation}
    \mathbf{z}_{\text{summary}} = \texttt{concat}\left( \mathbb{E}_{t}[\mathbf{z}_{t}], \mathbf{z}_{\frac{N}{4}}, \mathbf{z}_{\frac{N}{2}}, \mathbf{z}_{\frac{3N}{4}} \right),
\end{equation}
where $N$ is the sequence length.
We attach a 3-way linear shape classifier head to this vector.

\begin{figure}[h]
    \centering
    \includegraphics[width=\linewidth]{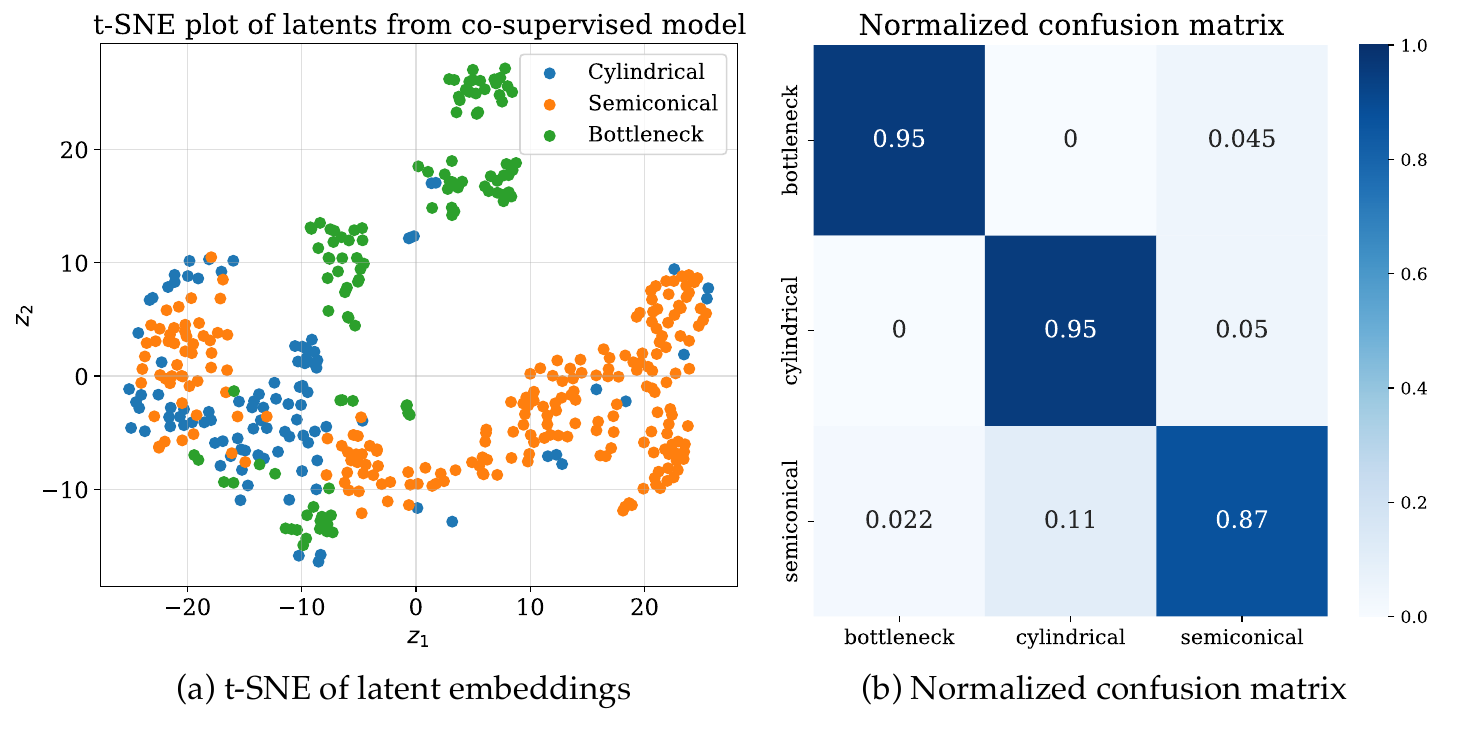}
    \captionsetup{font=normal,skip=1pt}
    \caption{\textbf{Results on shape recognition from audio embeddings.} (a) Latent embeddings learned by the co-supervised audio model encode container shape. (b) On an unseen test set, it is able to recognize the container shapes with a sample accuracy of 90.91\% and mean class accuracy of 92.47\%.
        Some semiconical containers with small difference between the top and base radius are essentially cylindrical and that shows in the t-SNE embeddings.
    }
    \label{fig:shape-prediction}
\end{figure}

To report performance, we consider the \textit{Test Set III} set of our dataset consisting of 434 videos (227 semiconical, 107 bottleneck and 100 cylindrical).
These are unseen samples not part of the training and evaluation set used previously to estimate physical properties.
We split these randomly in an 80-20 split.
On this split, we achieve a per sample accuracy of 90.91\% and a mean class accuracy of 92.47\%. t-SNE~\cite{van2008visualizing} embeddings and the normalized confusion matrix are shown in \cref{fig:shape-prediction}.
In comparison, features without co-supervision achieve 88.63\% and 89.44\%.
This shows: (i) a model trained to detect pitch implicitly encodes container shape, and (ii) co-supervision further improves shape recognition.

\subsection{Estimating liquid weight from its pouring sound}

The weight of the liquid being poured is a function of its density (constant) and the container volume it occupies. Here, we evaluate whether the learned audio representations can be regressed to predict the weight of the liquid. Since our dataset does not have weight annotations, we evaluate on the dataset proposed by \citet{Wilson2019AnalyzingLP}.

While \cite{Wilson2019AnalyzingLP} directly fine-tune on their dataset, we only linearly probe representations from our co-supervised network. Given input sound, we cut 0.4s snippets and compute a sequence of features $\{\mathbf{z}_{i}\}_{i=1}^{N}$. We attach a linear regressor head which outputs a scalar weight for each feature vector $\mathbf{z}_{i}$.

For transparent comparison, we report comparison with the same baselines, on the same data and splits as in \cite{Wilson2019AnalyzingLP}. Specifically, it consists of 136 video sequences with weight annotations across six different containers and two liquids. Following \cite{Wilson2019AnalyzingLP}, for each container, we train a separate regressor on its own split and report performance in \cref{tab:mass-est}. Our model outperforms all the baselines as well as the strong supervised baseline in \cite{Wilson2019AnalyzingLP}. On average, our model achieves an MAE of 1.20 oz which is the best amongst all baselines. Not surprisingly, our model performs the best with a cylindrical container (Column 4) but also is impressive on a bottleneck container (Columns 5 and 6). This shows that a network trained for pitch detection learns features capable of estimating liquid mass which is a function of the container volume and liquid density.
In \cref{appendix-subsec:other-results}, we also evaluate generalization across containers where we train a mass regressor on sounds of one container and evaluate it on those of another container.

\begin{table}[t!]
    \centering
    \resizebox{\columnwidth}{!}{%
        \begin{tabular}{lcccccccc}
            \toprule
            \textbf{Material}                                                      & \textbf{}     & Plastic                                                              & Glass           & Porcelain       & Metal           & Glass           & Glass                              \\
            \arrayrulecolor{lightgray}\cmidrule{1-8}
            \textbf{Shape}                                                         & \textbf{}     & Semiconical                                                          & Semiconical     & Pyramidal       & Cylindrical     & Bottle          & Bottle          &                  \\
            \arrayrulecolor{lightgray}\cmidrule{1-8}
            \textbf{Liquid}                                                        & \textbf{}     & Water                                                                & Water           & Water           & Water           & Milk            & Water           &                  \\
            \arrayrulecolor{lightgray}\cmidrule{1-8}

            \textbf{Example image}                                                 & \textbf{}     & {\includegraphics[width=0.1\textwidth]{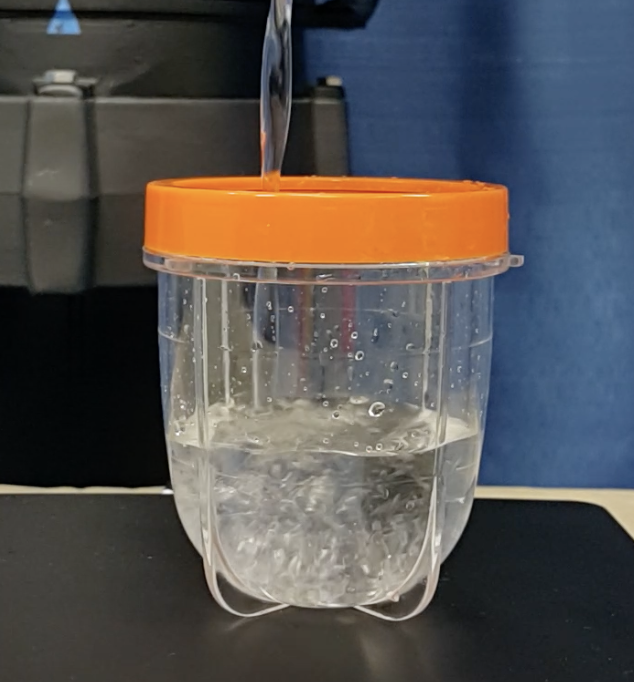}} &
            {\includegraphics[width=0.1\textwidth]{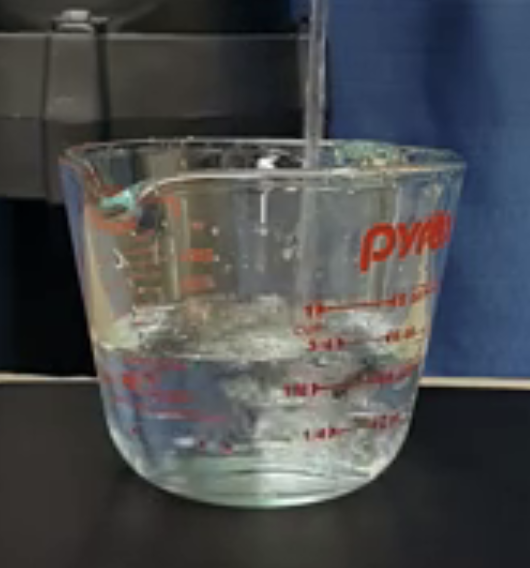}}     &
            {\includegraphics[width=0.1\textwidth]{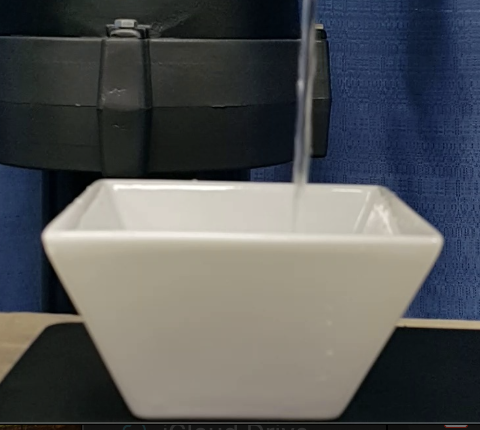}} &
            {\includegraphics[width=0.1\textwidth]{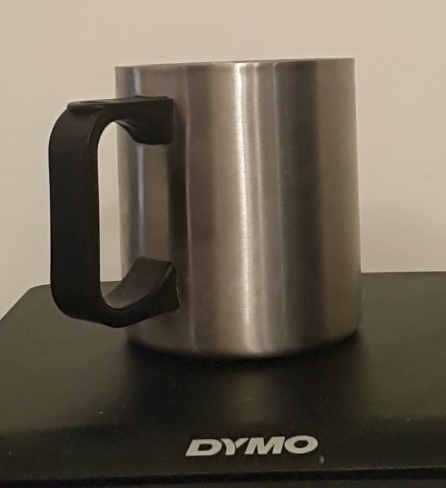}}     &
            {\includegraphics[width=0.1\textwidth]{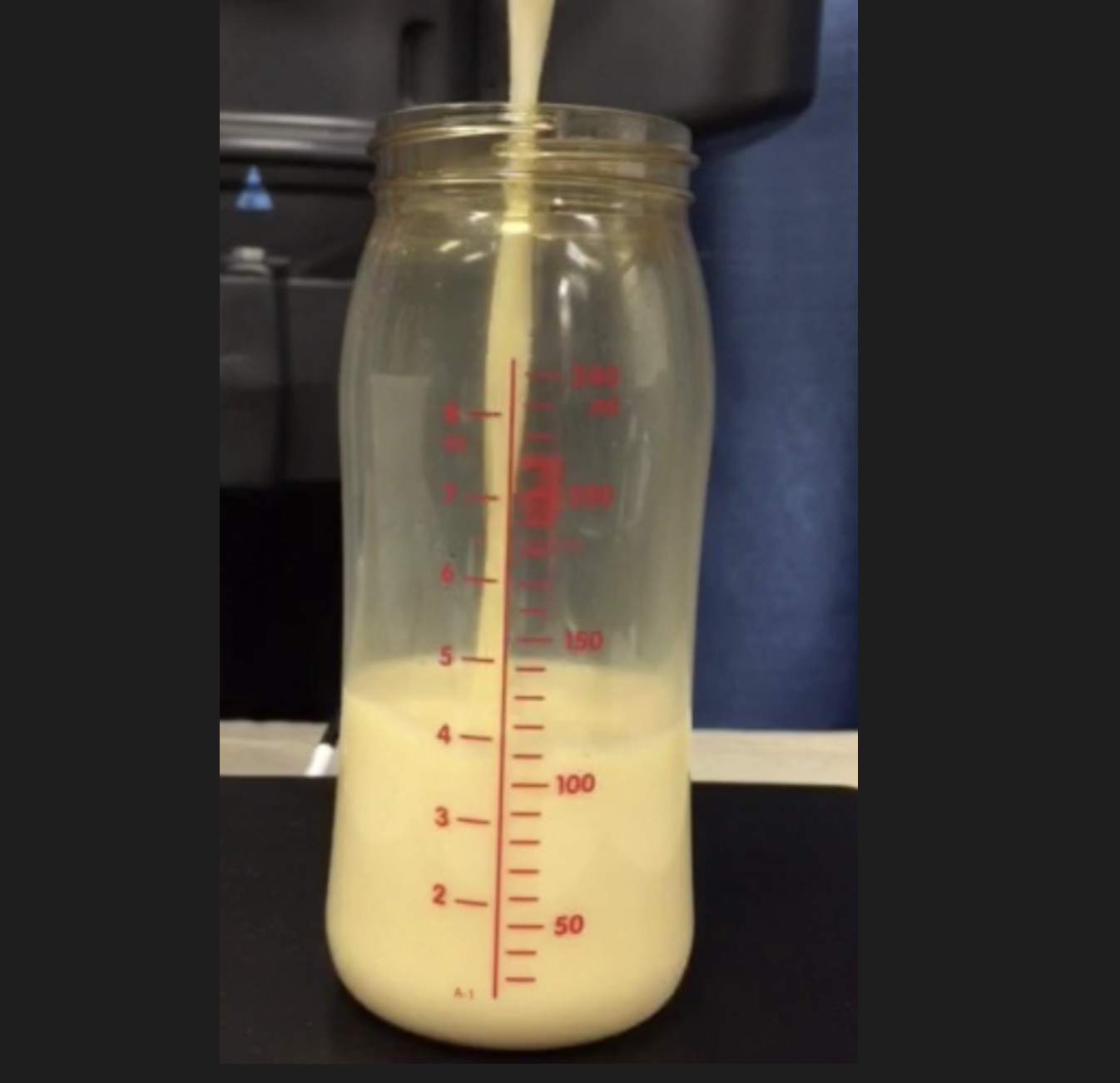}}   &
            {\includegraphics[width=0.1\textwidth]{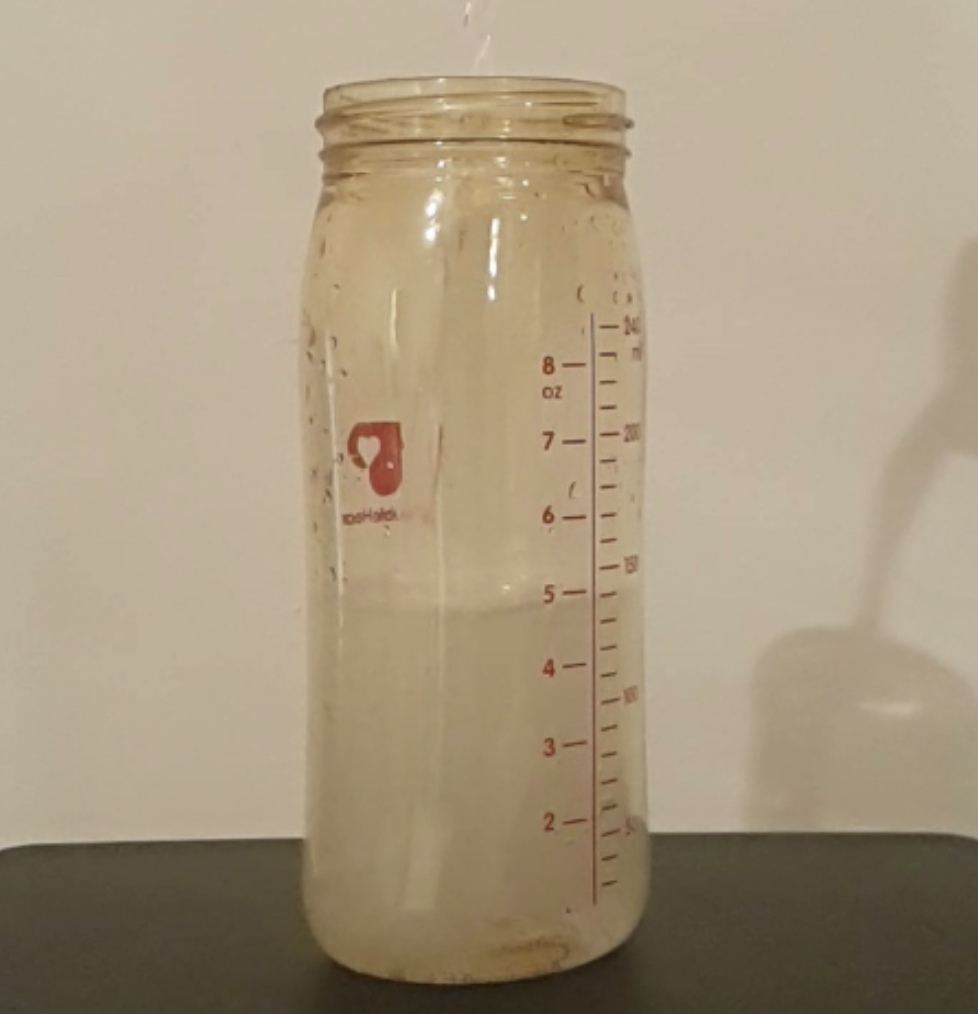}}  & \textbf{Mean}
            \\
            \arrayrulecolor{black}\midrule
            k-NN                                                                   &               & 3.4                                                                  & 3.6             & 2.2             & 2.5             & 2.7             & 2.7             & 2.85             \\
            Linear SVM                                                             &               & 3.4                                                                  & 4.8             & 3.3             & 4.1             & 3.5             & 4.3             & 3.90             \\
            SoundNet-5~(\citet{aytar2016soundnet})                                 &               & 3.4                                                                  & 4.2             & 4.4             & 4.7             & 3.0             & 3.6             & 3.88             \\
            SoundNet-8~(\citet{aytar2016soundnet})                                 &               & 3.2                                                                  & 6.1             & 3.5             & 4.2             & 5.8             & 4.7             & 4.58             \\
            TCN (\citet{lea2017temporal})                                          &               & 1.5                                                                  & 1.9             & 2.0             & 1.7             & 3.9             & 3.7             & 2.45             \\
            PSNN (\citet{Wilson2019AnalyzingLP})                                   &               & \textbf{1.2}                                                         & \textbf{1.2}    & \textbf{1.3}    & \textbf{0.7}    & \underline{1.8} & \underline{1.9} & \underline{1.35} \\
            Ours                                                                   &               & \underline{1.3}                                                      & \underline{1.3} & \underline{1.4} & \underline{0.9} & \textbf{1.0}    & \textbf{1.3}    & \textbf{1.20}    \\
            \arrayrulecolor{black}\bottomrule
        \end{tabular}
    }
    \caption{
        \textbf{Estimating liquid weight from its pouring sound.}
        Average MAE in mass estimation (in ounces, (oz)) from short
        snippets of the sound of pouring across different container
        and liquid configurations from the dataset by
        \citet{Wilson2019AnalyzingLP}. Linear probing our pre-trained
        co-supervised features outperforms all the baselines and
        performs competitively to the strong supervised fine-tuning
        baseline in \cite{Wilson2019AnalyzingLP}. Note that visual
        information is not used in any form. We pre-train on our
        dataset and linear probe on the evaluation dataset
        while \citet{Wilson2019AnalyzingLP} fine-tune on this dataset.
    }
    \label{tab:mass-est}
\end{table}

\subsection{Generalization and failure cases}

\begin{figure}[h]
    \centering
    \includegraphics[width=\textwidth]{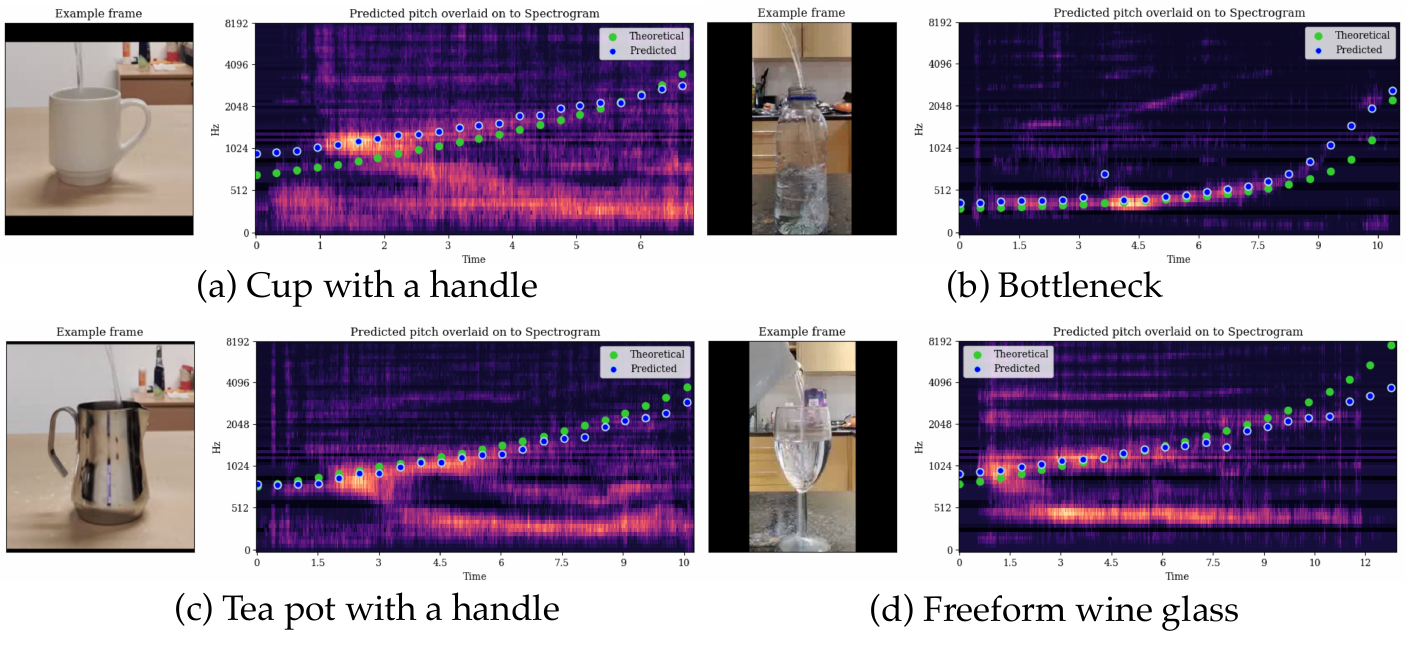}
    \captionsetup{font=normal,skip=0.5pt}
    \caption{\textbf{Generalization across container shapes.} Although our pitch predictor is trained only on cylinder-like containers, it works reasonably well on various free-form shapes encountered in daily use (unseen during training). The theoretical estimate (green curve) is obtained assuming a cylinder or a bottle and is an imperfect measure of the pitch. Nonetheless, the prediction (blue curve) tracks the pitch accurately even though it may disagree with the theoretical estimate.
    }
    \label{fig:gen-across-shapes}
\end{figure}

\begin{figure}[h]
    \centering
    \includegraphics[width=\textwidth]
    {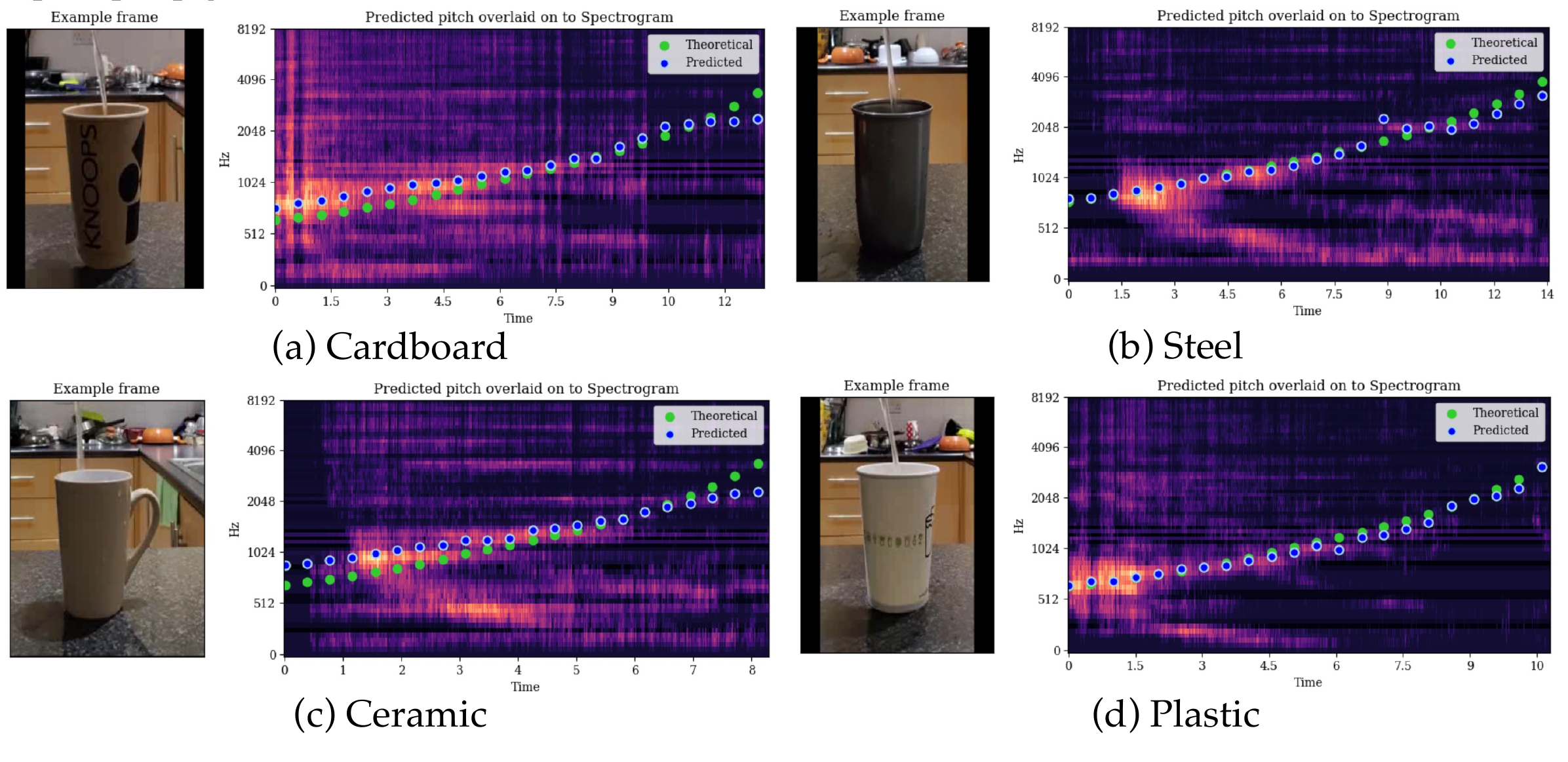}
    \captionsetup{font=normal,skip=0.5pt}
    \caption{\textbf{Generalization across container materials.} Our pitch detector works well for diverse kinds of container materials while the shape (semi-conical) is fixed. The theoretical estimate (green curve) is obtained by assuming it as a cylinder and thus is not perfect and only used for reference.}
    \label{fig:gen-across-materials}
\end{figure}

\begin{figure}[h]
    \centering
    \includegraphics[width=\textwidth]{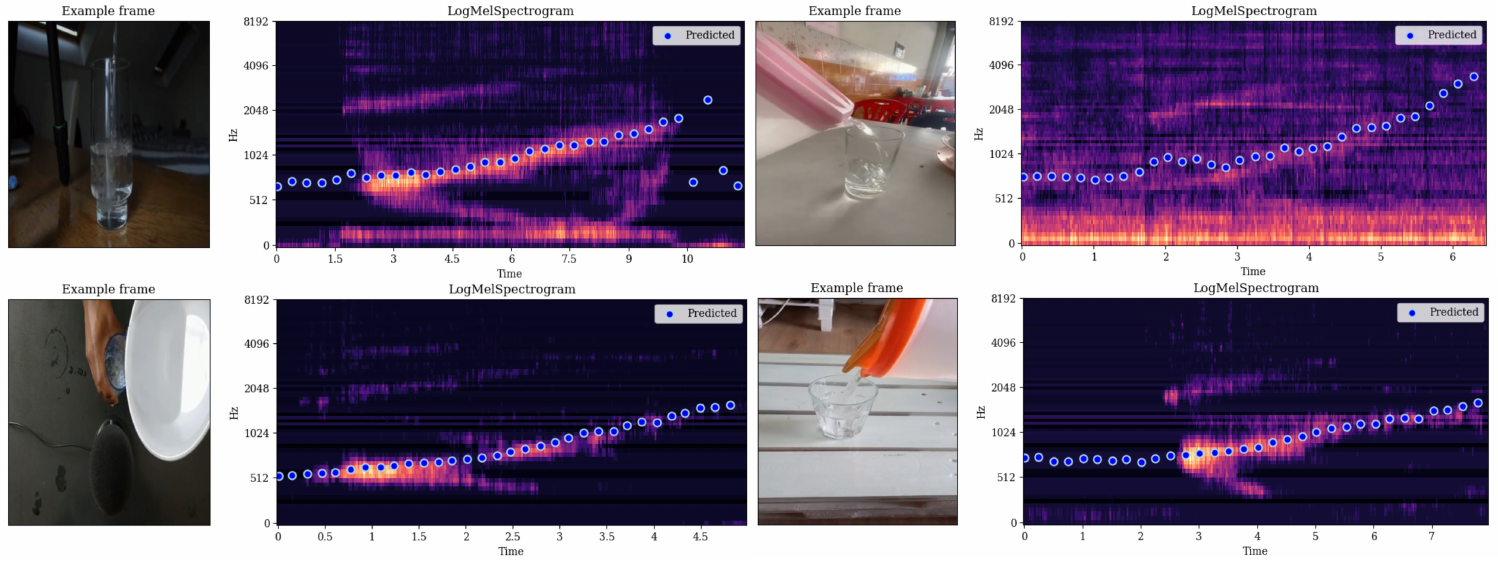}
    \captionsetup{font=normal,skip=0.8pt}
    \caption{\textbf{Generalization to in-the-wild videos.} We qualitatively evaluate on videos sourced from YouTube. Our pitch detector works very well even on these samples. Notice the variability in the visual inputs and contrast that to the consistency in the audio recordings.}
    \label{fig:youtube-results}
\end{figure}

\begin{figure}
    \centering
    \includegraphics[width=\linewidth]{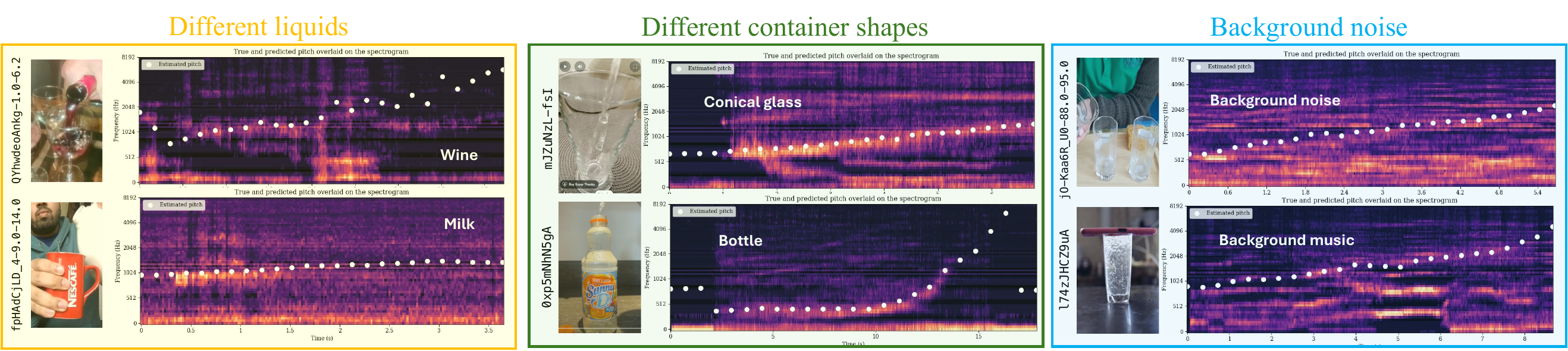}
    \captionsetup{font=normal,skip=4pt}
    \caption{\textbf{Robustness to various factors.} The model generalizes fairly well to variations in liquids, container shapes and to severe background noise. All samples are sourced from YouTube with IDs provided. We recommend the reader to try out such examples in the \href{https://huggingface.co/spaces/bpiyush/SoundOfWater}{online demo}.}
    \label{fig:robustness}
\end{figure}

While our pitch detection model is trained on cylinder-like transparent containers of glass and plastic in a clean setting, we qualitatively test if it generalizes to slightly different container shapes, container materials, and to samples from other datasets (e.g., \cite{Wilson2019AnalyzingLP}). To test generalization across shapes, we evaluate on various (unseen) containers of different shapes from our dataset (e.g., a cup or a teapot) and report robust results in \cref{fig:gen-across-shapes}. To test generalization across materials, we pick containers of the same shape (semi-conical) of various materials (e.g., steel, ceramic, cardboard, etc.) from our dataset. We find fairly convincing evidence for generalization across materials in \cref{fig:gen-across-materials}.
Finally, we also evaluate on samples in-the-wild from YouTube. As shown in \cref{fig:youtube-results}, the predictions fairly accurately track the fundamental frequency curve. Note the variability in the video which makes it hard to adjudge the container shape or size from vision, but it is much easier to infer from the detected pitch from out model. More examples of robustness to different liquids, container shapes and background noise is shown in \cref{fig:robustness}.

\begin{figure}
    \centering
    \includegraphics[width=\linewidth]{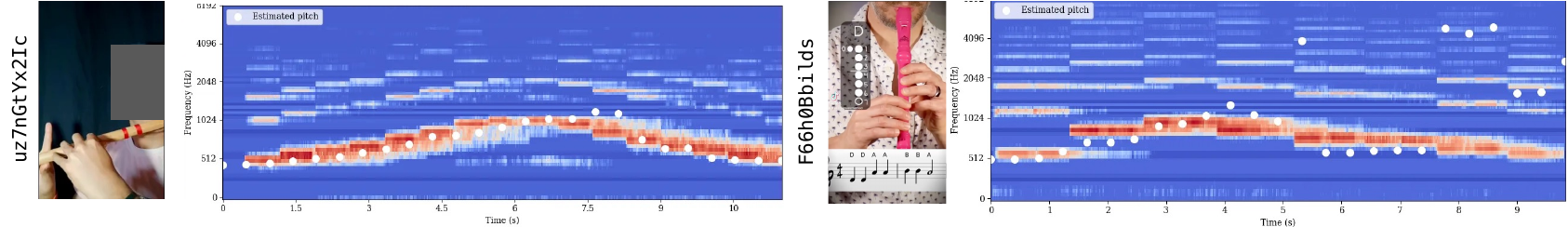}
    \captionsetup{font=normal,skip=1pt}
    \caption{\textbf{Generalization to music.} Qualitatively, we find surprisingly reasonable generalization to flute sounds likely since the same underlying physical phenomenon produces resonance. We do not find generalization to other musical instruments or polyphonic music due to the large domain gap. The YouTube IDs of the samples are provided alongside the image.}
    \label{fig:flutes}
\end{figure}

\paragraph{Generalization to music.}
While pitch detectors trained on music data do not generalize on pouring sounds (\cref{tab:pitch-detection-results}), it worth asking if the reverse is true. Given the large domain gap, we do not expect the model to work in an entirely different domain (like polyphonic music). However, as shown in \cref{fig:flutes}, it is slightly surprising that the pitch detector trained only on pouring sounds does reasonably well on flute sounds. This is likely because the same underlying principle that produces sound in both cases. It does not generalize to other musical instruments or polyphonic music.

\begin{figure}
    \centering
    \includegraphics[width=\linewidth]{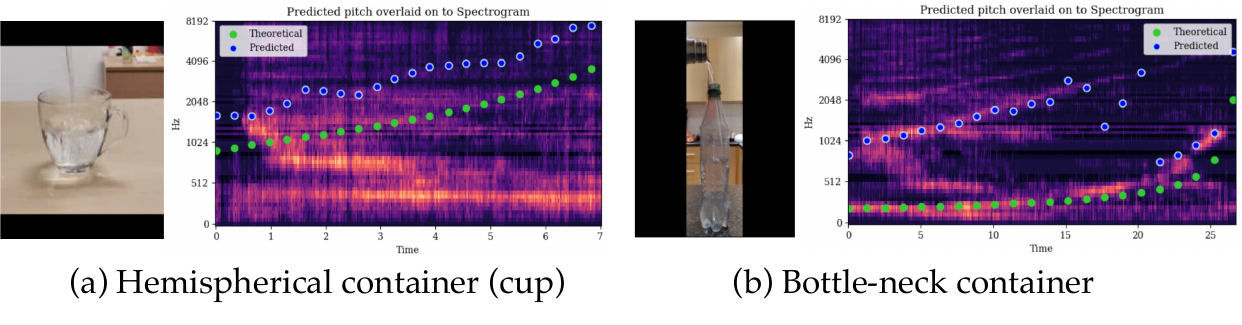}
    \captionsetup{font=normal,skip=1pt}
    \caption{\textbf{Failure cases.} (a) The model struggles in cases of hemispherical containers that have too much room for air to pass leading to very weak resonance (b) another challenging case is that of bottle-neck containers that show more than one mode in the frequency distribution on spectrogram.}
    \label{fig:failure-cases}
\end{figure}

\paragraph{Failure cases.}
Our pitch detector model struggles with hemispherical containers (e.g., a bowl as shown in \cref{fig:failure-cases} (b)). In such cases, there is a lot of room for air to escape which prevents the sort of ``filling-up'' effect seen in other containers.
It also struggles with certain cases where multiple frequency modes are present. For example, in some bottleneck containers (\cref{fig:failure-cases} (a)), along with the fundamental, a strong linear frequency curve is also present. The model fails to pick both together.

\section{Conclusions and extensions}
\label{sec:conclusion}

In this work, we considered the case of pouring water in a
container and analysed the connection between the underlying
physics and audio-visual observations. While humans can
reason well about the physical properties (e.g., absolute
container size) merely from the sound of pouring, we have shown
early evidence of training machines to achieve similar
capabilities. We developed synthetic data to pre-train an
audio network for pitch detection. We fine-tuned it on
real data with no external supervision thanks to the
co-supervision from the video stream. We demonstrated that
basic physical properties can be recovered accurately
from the estimated pitch. We showed that the co-supervised
representations also encode other useful properties such as
container shape and liquid weight. Finally, we showed promising
generalization to different container shapes, materials, and
in-the-wild videos, e.g., radial resonance as described
in \cref{subsec:radial}~\cite{Helmholtz2005OnTS, peng2018can}.
We hope that our work also prompts similar studies for physical
understanding from the sound of other mundane activities.

\paragraph{Acknowledgements.}
We thank Ashish Thandavan for support with infrastructure and
Sindhu Hegde, Ragav Sachdeva, Jaesung Huh, Vladimir Iashin,
Prajwal KR, and Aditya Singh  for useful discussions.
We also thank Justin Wilson for their help with some data
details from \citet{Wilson2019AnalyzingLP}. We also thank anonymous reviewers that helped improve this work.
This research is funded by EPSRC Programme Grant VisualAI
EP/T028572/1, and a Royal Society Research Professorship RP / R1 / 191132.

\newpage
{
  \footnotesize
  \bibliographystyle{plainnat}
  \bibliography{longstrings,ref}

\begin{thebibliography}{101}
\providecommand{\natexlab}[1]{#1}
\providecommand{\url}[1]{\texttt{#1}}
\expandafter\ifx\csname urlstyle\endcsname\relax
  \providecommand{\doi}[1]{doi: #1}\else
  \providecommand{\doi}{doi: \begingroup \urlstyle{rm}\Url}\fi

\bibitem[Afouras et~al.(2018{\natexlab{a}})Afouras, Chung, and
  Zisserman]{Afouras18b}
T.~Afouras, J.~S. Chung, and A.~Zisserman.
\newblock Deep lip reading: a comparison of models and an online application.
\newblock In \emph{Conference of the International Speech Communication
  Association (INTERSPEECH)}, 2018{\natexlab{a}}.

\bibitem[Afouras et~al.(2018{\natexlab{b}})Afouras, Chung, Senior, Vinyals, and
  Zisserman]{afouras2018deep}
Triantafyllos Afouras, Joon~Son Chung, Andrew Senior, Oriol Vinyals, and Andrew
  Zisserman.
\newblock Deep audio-visual speech recognition.
\newblock \emph{IEEE Transactions on Pattern Analysis and Machine Intelligence
  (TPAMI)}, 44\penalty0 (12):\penalty0 8717--8727, 2018{\natexlab{b}}.

\bibitem[Agrawal et~al.(2020)Agrawal, Lee, Calcetas, Clarke, Lin, and
  Schachner]{temperature2_Agrawal2020HearingWT}
Tanushree Agrawal, Michelle Lee, Amanda Calcetas, Danielle Clarke, Naomi Lin,
  and Adena Schachner.
\newblock Hearing water temperature: Characterizing the development of nuanced
  perception of auditory events.
\newblock In \emph{Annual Meeting of the Cognitive Science Society}, 2020.
\newblock URL \url{https://api.semanticscholar.org/CorpusID:231792766}.

\bibitem[Akbari et~al.(2021)Akbari, Yuan, Qian, Chuang, Chang, Cui, and
  Gong]{akbari2021vatt}
Hassan Akbari, Liangzhe Yuan, Rui Qian, Wei-Hong Chuang, Shih-Fu Chang, Yin
  Cui, and Boqing Gong.
\newblock Vatt: Transformers for multimodal self-supervised learning from raw
  video, audio and text.
\newblock \emph{Advances in Neural Information Processing Systems (NeurIPS)},
  34:\penalty0 24206--24221, 2021.

\bibitem[Alwassel et~al.(2020)Alwassel, Mahajan, Korbar, Torresani, Ghanem, and
  Tran]{alwassel2020self}
Humam Alwassel, Dhruv Mahajan, Bruno Korbar, Lorenzo Torresani, Bernard Ghanem,
  and Du~Tran.
\newblock Self-supervised learning by cross-modal audio-video clustering.
\newblock \emph{Advances in Neural Information Processing Systems (NeurIPS)},
  33:\penalty0 9758--9770, 2020.

\bibitem[Anderson and Ostensen(1928)]{Anderson1928EffectOF}
S.~Herbert Anderson and Floyd~C. Ostensen.
\newblock Effect of frequency on the end correction of pipes.
\newblock \emph{Physical Review}, 1928.
\newblock URL \url{https://api.semanticscholar.org/CorpusID:121139086}.

\bibitem[Arandjelovic and Zisserman(2017)]{arandjelovic2017look}
Relja Arandjelovic and Andrew Zisserman.
\newblock Look, listen and learn.
\newblock In \emph{International Conference on Computer Vision (ICCV)}, pages
  609--617, 2017.

\bibitem[Arandjelovic and Zisserman(2018)]{arandjelovic2018objects}
Relja Arandjelovic and Andrew Zisserman.
\newblock Objects that sound.
\newblock In \emph{European Conference on Computer Vision (ECCV)}, pages
  435--451, 2018.

\bibitem[Asano et~al.(2020)Asano, Patrick, Rupprecht, and
  Vedaldi]{asano2020labelling}
Yuki Asano, Mandela Patrick, Christian Rupprecht, and Andrea Vedaldi.
\newblock Labelling unlabelled videos from scratch with multi-modal
  self-supervision.
\newblock \emph{Advances in Neural Information Processing Systems (NeurIPS)},
  33:\penalty0 4660--4671, 2020.

\bibitem[Aytar et~al.(2016)Aytar, Vondrick, and Torralba]{aytar2016soundnet}
Yusuf Aytar, Carl Vondrick, and Antonio Torralba.
\newblock Soundnet: Learning sound representations from unlabeled video.
\newblock \emph{Advances in Neural Information Processing Systems (NeurIPS)},
  29, 2016.

\bibitem[Babaians et~al.(2022)Babaians, Sharma, Karimi, Sharifzadeh, and
  Steinbach]{pouring7_9981195}
Edwin Babaians, Tapan Sharma, Mojtaba Karimi, Sahand Sharifzadeh, and Eckehard
  Steinbach.
\newblock Pournet: Robust robotic pouring through curriculum and
  curiosity-based reinforcement learning.
\newblock In \emph{IEEE/RSJ International Conference on Intelligent Robots and
  Systems (IROS)}, pages 9332--9339, 2022.
\newblock \doi{10.1109/IROS47612.2022.9981195}.

\bibitem[Baevski et~al.(2020)Baevski, Zhou, Mohamed, and
  Auli]{baevski2020wav2vec}
Alexei Baevski, Yuhao Zhou, Abdelrahman Mohamed, and Michael Auli.
\newblock wav2vec 2.0: A framework for self-supervised learning of speech
  representations.
\newblock \emph{Advances in Neural Information Processing Systems (NeurIPS)},
  33:\penalty0 12449--12460, 2020.

\bibitem[Bagad et~al.(2025)Bagad, Tapaswi, Snoek, and
  Zisserman]{bagad2024soundofwater}
Piyush Bagad, Makarand Tapaswi, Cees G.~M. Snoek, and Andrew Zisserman.
\newblock The {S}ound of {W}ater: {I}nferring {P}hysical {P}roperties from
  {P}ouring {L}iquids.
\newblock In \emph{ICASSP}, 2025.

\bibitem[Banerji(1919)]{radial2-PhysRev.13.171}
Sudhansukumar Banerji.
\newblock On the vibrations of elastic shells partly filled with liquid.
\newblock \emph{Phys. Rev.}, Mar 1919.
\newblock \doi{10.1103/PhysRev.13.171}.
\newblock URL \url{https://link.aps.org/doi/10.1103/PhysRev.13.171}.

\bibitem[Berg and Stork(1982)]{berg1982physics}
Richard~E Berg and David~G Stork.
\newblock \emph{The physics of sound}.
\newblock Pearson Education India, 1982.

\bibitem[Cabe and Pittenger(2000)]{Cabe2000HumanST}
Patrick~A. Cabe and John~B. Pittenger.
\newblock Human sensitivity to acoustic information from vessel filling.
\newblock \emph{Journal of experimental psychology. Human perception and
  performance}, 2000.

\bibitem[Carello et~al.(1998)Carello, Anderson, and
  Kunkler-Peck]{carello1998perception}
Claudia Carello, Krista~L Anderson, and Andrew~J Kunkler-Peck.
\newblock Perception of object length by sound.
\newblock \emph{Psychological science}, 9\penalty0 (3):\penalty0 211--214,
  1998.

\bibitem[Caron et~al.(2021)Caron, Touvron, Misra, J{\'e}gou, Mairal,
  Bojanowski, and Joulin]{caron2021emerging}
Mathilde Caron, Hugo Touvron, Ishan Misra, Herv{\'e} J{\'e}gou, Julien Mairal,
  Piotr Bojanowski, and Armand Joulin.
\newblock Emerging properties in self-supervised vision transformers.
\newblock In \emph{International Conference on Computer Vision (ICCV)}, pages
  9650--9660, 2021.

\bibitem[Chen et~al.(2021{\natexlab{a}})Chen, Rouditchenko, Duarte, Kuehne,
  Thomas, Boggust, Panda, Kingsbury, Feris, Harwath,
  et~al.]{chen2021multimodal}
Brian Chen, Andrew Rouditchenko, Kevin Duarte, Hilde Kuehne, Samuel Thomas,
  Angie Boggust, Rameswar Panda, Brian Kingsbury, Rogerio Feris, David Harwath,
  et~al.
\newblock Multimodal clustering networks for self-supervised learning from
  unlabeled videos.
\newblock In \emph{International Conference on Computer Vision (ICCV)}, pages
  8012--8021, 2021{\natexlab{a}}.

\bibitem[Chen et~al.(2020)Chen, Xie, Vedaldi, and Zisserman]{vggsound}
Honglie Chen, Weidi Xie, Andrea Vedaldi, and Andrew Zisserman.
\newblock Vggsound: A large-scale audio-visual dataset.
\newblock In \emph{International Conference on Acoustics, Speech and Signal
  Processing (ICASSP)}, 2020.

\bibitem[Chen et~al.(2021{\natexlab{b}})Chen, Xie, Afouras, Nagrani, Vedaldi,
  and Zisserman]{chen2021localizing}
Honglie Chen, Weidi Xie, Triantafyllos Afouras, Arsha Nagrani, Andrea Vedaldi,
  and Andrew Zisserman.
\newblock Localizing visual sounds the hard way.
\newblock In \emph{Conference on Computer Vision and Pattern Recognition
  (CVPR)}, pages 16867--16876, 2021{\natexlab{b}}.

\bibitem[Chen et~al.(2019)Chen, Huang, and Sun]{pouring5_8967802}
Tianze Chen, Yongqiang Huang, and Yu~Sun.
\newblock Accurate pouring using model predictive control enabled by recurrent
  neural network.
\newblock In \emph{IEEE/RSJ International Conference on Intelligent Robots and
  Systems (IROS)}, pages 7688--7694, 2019.
\newblock \doi{10.1109/IROS40897.2019.8967802}.

\bibitem[Chen et~al.(2022)Chen, Fouhey, and Owens]{chen2022sound}
Ziyang Chen, David~F Fouhey, and Andrew Owens.
\newblock Sound localization by self-supervised time delay estimation.
\newblock In \emph{European Conference on Computer Vision (ECCV)}, pages
  489--508. Springer, 2022.

\bibitem[Chung et~al.(2024)Chung, Lee, and Nam]{chung2024t}
Yoonjin Chung, Junwon Lee, and Juhan Nam.
\newblock T-foley: A controllable waveform-domain diffusion model for
  temporal-event-guided foley sound synthesis.
\newblock In \emph{International Conference on Acoustics, Speech and Signal
  Processing (ICASSP)}, pages 6820--6824. IEEE, 2024.

\bibitem[Damen et~al.(2022)Damen, Doughty, Farinella, Furnari, Ma, Kazakos,
  Moltisanti, Munro, Perrett, Price, and Wray]{Damen2022RESCALING}
Dima Damen, Hazel Doughty, Giovanni~Maria Farinella, Antonino Furnari, Jian Ma,
  Evangelos Kazakos, Davide Moltisanti, Jonathan Munro, Toby Perrett, Will
  Price, and Michael Wray.
\newblock Rescaling egocentric vision: Collection, pipeline and challenges for
  epic-kitchens-100.
\newblock \emph{International Journal of Computer Vision (IJCV)}, 130:\penalty0
  33–55, 2022.
\newblock URL \url{https://doi.org/10.1007/s11263-021-01531-2}.

\bibitem[De~Cheveign{\'e} and Kawahara(2002)]{de2002yin}
Alain De~Cheveign{\'e} and Hideki Kawahara.
\newblock Yin, a fundamental frequency estimator for speech and music.
\newblock \emph{The Journal of the Acoustical Society of America}, 111\penalty0
  (4):\penalty0 1917--1930, 2002.

\bibitem[Do et~al.(2016)Do, Schubert, and Burgard]{liquid_perception2_7759326}
Chau Do, Tobias Schubert, and Wolfram Burgard.
\newblock A probabilistic approach to liquid level detection in cups using an
  rgb-d camera.
\newblock In \emph{IEEE/RSJ International Conference on Intelligent Robots and
  Systems (IROS)}, pages 2075--2080, 2016.
\newblock \doi{10.1109/IROS.2016.7759326}.

\bibitem[Do et~al.(2018)Do, Gordillo, and Burgard]{pouring8_8593654}
Chau Do, Camilo Gordillo, and Wolfram Burgard.
\newblock Learning to pour using deep deterministic policy gradients.
\newblock In \emph{IEEE/RSJ International Conference on Intelligent Robots and
  Systems (IROS)}, pages 3074--3079, 2018.
\newblock \doi{10.1109/IROS.2018.8593654}.

\bibitem[Dong et~al.(2019)Dong, Takizawa, Kudoh, and Suehiro]{pourin6_8967911}
Chenyu Dong, Masaru Takizawa, Shunsuke Kudoh, and Takashi Suehiro.
\newblock Precision pouring into unknown containers by service robots.
\newblock In \emph{IEEE/RSJ International Conference on Intelligent Robots and
  Systems (IROS)}, pages 5875--5882, 2019.
\newblock \doi{10.1109/IROS40897.2019.8967911}.

\bibitem[Du et~al.(2023)Du, Chen, Salamon, Russell, and
  Owens]{du2023conditional}
Yuexi Du, Ziyang Chen, Justin Salamon, Bryan Russell, and Andrew Owens.
\newblock Conditional generation of audio from video via foley analogies.
\newblock In \emph{Conference on Computer Vision and Pattern Recognition
  (CVPR)}, pages 2426--2436, 2023.

\bibitem[Engel et~al.(2020)Engel, Hantrakul, Gu, and Roberts]{engel2020ddsp}
Jesse Engel, Lamtharn~(Hanoi) Hantrakul, Chenjie Gu, and Adam Roberts.
\newblock Ddsp: Differentiable digital signal processing.
\newblock In \emph{International Conference on Learning Representations
  (ICLR)}, 2020.
\newblock URL \url{https://openreview.net/forum?id=B1x1ma4tDr}.

\bibitem[Eppel et~al.(2021)Eppel, Xu, Wang, and
  Aspuru-Guzik]{liquid_perception6_eppel2021predicting}
Sagi Eppel, Haoping Xu, Yi~Ru Wang, and Alan Aspuru-Guzik.
\newblock Predicting 3d shapes, masks, and properties of materials, liquids,
  and objects inside transparent containers, using the transproteus cgi
  dataset, 2021.

\bibitem[French(1983)]{radial1-French1983InVV}
Anthony~P. French.
\newblock In vino veritas: A study of wineglass acoustics.
\newblock \emph{American Journal of Physics}, 1983.
\newblock URL \url{https://api.semanticscholar.org/CorpusID:120875058}.

\bibitem[Gemmeke et~al.(2017)Gemmeke, Ellis, Freedman, Jansen, Lawrence, Moore,
  Plakal, and Ritter]{audioset}
Jort~F. Gemmeke, Daniel P.~W. Ellis, Dylan Freedman, Aren Jansen, Wade
  Lawrence, R.~Channing Moore, Manoj Plakal, and Marvin Ritter.
\newblock Audio set: An ontology and human-labeled dataset for audio events.
\newblock In \emph{International Conference on Acoustics, Speech and Signal
  Processing (ICASSP)}, 2017.

\bibitem[Georgescu et~al.(2023)Georgescu, Fonseca, Ionescu, Lucic, Schmid, and
  Arnab]{georgescu2023audiovisual}
Mariana-Iuliana Georgescu, Eduardo Fonseca, Radu~Tudor Ionescu, Mario Lucic,
  Cordelia Schmid, and Anurag Arnab.
\newblock Audiovisual masked autoencoders.
\newblock In \emph{International Conference on Computer Vision (ICCV)}, pages
  16144--16154, 2023.

\bibitem[Gibson(2014)]{gibson2014ecological}
James~J Gibson.
\newblock \emph{The ecological approach to visual perception: classic edition}.
\newblock Psychology press, 2014.

\bibitem[Girdhar et~al.(2023)Girdhar, El-Nouby, Liu, Singh, Alwala, Joulin, and
  Misra]{girdhar2023imagebind}
Rohit Girdhar, Alaaeldin El-Nouby, Zhuang Liu, Mannat Singh, Kalyan~Vasudev
  Alwala, Armand Joulin, and Ishan Misra.
\newblock Imagebind: One embedding space to bind them all.
\newblock In \emph{Conference on Computer Vision and Pattern Recognition
  (CVPR)}, pages 15180--15190, 2023.

\bibitem[Gong et~al.(2022)Gong, Rouditchenko, Liu, Harwath, Karlinsky, Kuehne,
  and Glass]{gong2022contrastive}
Yuan Gong, Andrew Rouditchenko, Alexander~H Liu, David Harwath, Leonid
  Karlinsky, Hilde Kuehne, and James Glass.
\newblock Contrastive audio-visual masked autoencoder.
\newblock \emph{arXiv:2210.07839}, 2022.

\bibitem[Gong et~al.(2023)Gong, Luo, Liu, Karlinsky, and Glass]{gong2023listen}
Yuan Gong, Hongyin Luo, Alexander~H Liu, Leonid Karlinsky, and James Glass.
\newblock Listen, think, and understand.
\newblock \emph{arXiv:2305.10790}, 2023.

\bibitem[Gordon et~al.(2013)Gordon, Russo, and MacDonald]{gordon2013spectral}
Michael~S Gordon, Frank~A Russo, and Ewen MacDonald.
\newblock Spectral information for detection of acoustic time to arrival.
\newblock \emph{Attention, Perception, \& Psychophysics}, 75:\penalty0
  738--750, 2013.

\bibitem[Guignard(2003)]{Guignard2003TuningOM}
Thomas Guignard.
\newblock Tuning of musical glasses.
\newblock In \emph{Master's Thesis, ETH Zurich}, 2003.
\newblock URL \url{https://api.semanticscholar.org/CorpusID:137983171}.

\bibitem[Guzhov et~al.(2022)Guzhov, Raue, Hees, and
  Dengel]{guzhov2022audioclip}
Andrey Guzhov, Federico Raue, J{\"o}rn Hees, and Andreas Dengel.
\newblock Audioclip: Extending clip to image, text and audio.
\newblock In \emph{International Conference on Acoustics, Speech and Signal
  Processing (ICASSP)}, pages 976--980. IEEE, 2022.

\bibitem[Helmholtz and Ellis(2005)]{Helmholtz2005OnTS}
Hermann L.~F. Helmholtz and Alexander~John Ellis.
\newblock On the sensations of tone as a physiological basis for the theory of
  music.
\newblock \emph{Nature}, 12:\penalty0 449--452, 2005.
\newblock URL \url{https://api.semanticscholar.org/CorpusID:119511156}.

\bibitem[Hendrycks and Gimpel(2016)]{hendrycks2016gaussian}
Dan Hendrycks and Kevin Gimpel.
\newblock Gaussian error linear units (gelus).
\newblock \emph{arXiv:1606.08415}, 2016.

\bibitem[Huang et~al.(2024)Huang, Sharma, Xu, Ryali, Li, Li, Ghosh, Malik,
  Feichtenhofer, et~al.]{huang2024mavil}
Po-Yao Huang, Vasu Sharma, Hu~Xu, Chaitanya Ryali, Yanghao Li, Shang-Wen Li,
  Gargi Ghosh, Jitendra Malik, Christoph Feichtenhofer, et~al.
\newblock Mavil: Masked audio-video learners.
\newblock \emph{Advances in Neural Information Processing Systems (NeurIPS)},
  36, 2024.

\bibitem[Huang and Sun(2017)]{pouring4_huang2017learning}
Yongqiang Huang and Yu~Sun.
\newblock Learning to pour, 2017.

\bibitem[Huang et~al.(2021)Huang, Wilches, and Sun]{pouring3_HUANG2021103692}
Yongqiang Huang, Juan Wilches, and Yu~Sun.
\newblock Robot gaining accurate pouring skills through self-supervised
  learning and generalization.
\newblock \emph{Robotics and Autonomous Systems}, 136:\penalty0 103692, 2021.

\bibitem[Jones(1941)]{Jones1941EndCO}
Arthur~Taber Jones.
\newblock End corrections of organ pipes.
\newblock \emph{Journal of the Acoustical Society of America}, 1941.
\newblock URL \url{https://api.semanticscholar.org/CorpusID:120564889}.

\bibitem[Kay et~al.(2017)Kay, Carreira, Simonyan, Zhang, Hillier,
  Vijayanarasimhan, Viola, Green, Back, Natsev, et~al.]{kay2017kinetics}
Will Kay, Joao Carreira, Karen Simonyan, Brian Zhang, Chloe Hillier, Sudheendra
  Vijayanarasimhan, Fabio Viola, Tim Green, Trevor Back, Paul Natsev, et~al.
\newblock The kinetics human action video dataset.
\newblock \emph{arXiv:1705.06950}, 2017.

\bibitem[Kim et~al.(2018)Kim, Salamon, Li, and Bello]{kim2018crepe}
Jong~Wook Kim, Justin Salamon, Peter Li, and Juan~Pablo Bello.
\newblock Crepe: A convolutional representation for pitch estimation.
\newblock In \emph{International Conference on Acoustics, Speech and Signal
  Processing (ICASSP)}, pages 161--165. IEEE, 2018.

\bibitem[Kingma(2014)]{kingma2014adam}
DP~Kingma.
\newblock Adam: a method for stochastic optimization.
\newblock \emph{arXiv:1412.6980}, 2014.

\bibitem[Kirillov et~al.(2023)Kirillov, Mintun, Ravi, Mao, Rolland, Gustafson,
  Xiao, Whitehead, Berg, Lo, et~al.]{kirillov2023segment}
Alexander Kirillov, Eric Mintun, Nikhila Ravi, Hanzi Mao, Chloe Rolland, Laura
  Gustafson, Tete Xiao, Spencer Whitehead, Alexander~C Berg, Wan-Yen Lo, et~al.
\newblock Segment anything.
\newblock In \emph{International Conference on Computer Vision (ICCV)}, pages
  4015--4026, 2023.

\bibitem[Klatzky et~al.(2000)Klatzky, Pai, and Krotkov]{klatzky2000perception}
Roberta~L Klatzky, Dinesh~K Pai, and Eric~P Krotkov.
\newblock Perception of material from contact sounds.
\newblock \emph{Presence}, 9\penalty0 (4):\penalty0 399--410, 2000.

\bibitem[Korbar et~al.(2018)Korbar, Tran, and Torresani]{korbar2018cooperative}
Bruno Korbar, Du~Tran, and Lorenzo Torresani.
\newblock Cooperative learning of audio and video models from self-supervised
  synchronization.
\newblock \emph{Advances in Neural Information Processing Systems (NeurIPS)},
  31, 2018.

\bibitem[Kunkler-Peck and Turvey(2000)]{kunkler2000hearing}
Andrew~J Kunkler-Peck and Michael~T Turvey.
\newblock Hearing shape.
\newblock \emph{Journal of Experimental psychology: human perception and
  performance}, 26\penalty0 (1):\penalty0 279, 2000.

\bibitem[Lea et~al.(2017)Lea, Flynn, Vidal, Reiter, and Hager]{lea2017temporal}
Colin Lea, Michael~D Flynn, Rene Vidal, Austin Reiter, and Gregory~D Hager.
\newblock Temporal convolutional networks for action segmentation and
  detection.
\newblock In \emph{Conference on Computer Vision and Pattern Recognition
  (CVPR)}, pages 156--165, 2017.

\bibitem[Lee et~al.(2022)Lee, Oh, Byeon, Kim, Ryoo, Yoon, Cho, Bae, Kim, and
  Kim]{lee2022sound}
Seung~Hyun Lee, Gyeongrok Oh, Wonmin Byeon, Chanyoung Kim, Won~Jeong Ryoo,
  Sang~Ho Yoon, Hyunjun Cho, Jihyun Bae, Jinkyu Kim, and Sangpil Kim.
\newblock Sound-guided semantic video generation.
\newblock In \emph{European Conference on Computer Vision (ECCV)}, pages
  34--50. Springer, 2022.

\bibitem[Liang et~al.(2019{\natexlab{a}})Liang, Li, Ma, Hendrich, Gerkmann,
  Sun, and Zhang]{Liang_2019}
Hongzhuo Liang, Shuang Li, Xiaojian Ma, Norman Hendrich, Timo Gerkmann, Fuchun
  Sun, and Jianwei Zhang.
\newblock Making sense of audio vibration for liquid height estimation in
  robotic pouring.
\newblock In \emph{IEEE/RSJ International Conference on Intelligent Robots and
  Systems (IROS)}, November 2019{\natexlab{a}}.

\bibitem[Liang et~al.(2019{\natexlab{b}})Liang, Li, Ma, Hendrich, Gerkmann,
  Sun, and Zhang]{liang2019making}
Hongzhuo Liang, Shuang Li, Xiaojian Ma, Norman Hendrich, Timo Gerkmann, Fuchun
  Sun, and Jianwei Zhang.
\newblock Making sense of audio vibration for liquid height estimation in
  robotic pouring.
\newblock In \emph{IEEE/RSJ International Conference on Intelligent Robots and
  Systems (IROS)}, pages 5333--5339. IEEE, 2019{\natexlab{b}}.

\bibitem[Liang et~al.(2020{\natexlab{a}})Liang, Zhou, Li, Ma, Hendrich,
  Gerkmann, Sun, Stoffel, and Zhang]{Liang_2020}
Hongzhuo Liang, Chuangchuang Zhou, Shuang Li, Xiaojian Ma, Norman Hendrich,
  Timo Gerkmann, Fuchun Sun, Marcus Stoffel, and Jianwei Zhang.
\newblock Robust robotic pouring using audition and haptics.
\newblock In \emph{IEEE/RSJ International Conference on Intelligent Robots and
  Systems (IROS)}. IEEE, October 2020{\natexlab{a}}.

\bibitem[Liang et~al.(2020{\natexlab{b}})Liang, Zhou, Li, Ma, Hendrich,
  Gerkmann, Sun, Stoffel, and Zhang]{liang2020robust}
Hongzhuo Liang, Chuangchuang Zhou, Shuang Li, Xiaojian Ma, Norman Hendrich,
  Timo Gerkmann, Fuchun Sun, Marcus Stoffel, and Jianwei Zhang.
\newblock Robust robotic pouring using audition and haptics.
\newblock In \emph{IEEE/RSJ International Conference on Intelligent Robots and
  Systems (IROS)}, pages 10880--10887. IEEE, 2020{\natexlab{b}}.

\bibitem[Lin et~al.(2023)Lin, Fu, and Xue]{liquid_perception5_Lin_2023_ICCV}
Haitao Lin, Yanwei Fu, and Xiangyang Xue.
\newblock Pourit!: Weakly-supervised liquid perception from a single image for
  visual closed-loop robotic pouring.
\newblock In \emph{International Conference on Computer Vision (ICCV)}, pages
  241--251, October 2023.

\bibitem[Liu et~al.(2020)Liu, Feng, Lan, and Chan]{Liu2020VA2MassTT}
Qi~Liu, Fan Feng, Chuanlin Lan, and Rosa H.~M. Chan.
\newblock Va2mass: Towards the fluid filling mass estimation via integration of
  vision and audio learning.
\newblock In \emph{ICPR Workshops}, 2020.
\newblock URL \url{https://api.semanticscholar.org/CorpusID:232023100}.

\bibitem[Ma et~al.(2020)Ma, Zeng, McDuff, and Song]{ma2020active}
Shuang Ma, Zhaoyang Zeng, Daniel McDuff, and Yale Song.
\newblock Active contrastive learning of audio-visual video representations.
\newblock \emph{arXiv:2009.09805}, 2020.

\bibitem[Medeiros(2024)]{langsam}
Luca Medeiros.
\newblock Language segment-anything.
\newblock Github, 2024.

\bibitem[Mershon and Bowers(1979)]{mershon1979absolute}
Donald~H Mershon and John~N Bowers.
\newblock Absolute and relative cues for the auditory perception of egocentric
  distance.
\newblock \emph{Perception}, 8\penalty0 (3):\penalty0 311--322, 1979.

\bibitem[Morgado et~al.(2021)Morgado, Vasconcelos, and Misra]{morgado2021audio}
Pedro Morgado, Nuno Vasconcelos, and Ishan Misra.
\newblock Audio-visual instance discrimination with cross-modal agreement.
\newblock In \emph{Conference on Computer Vision and Pattern Recognition
  (CVPR)}, pages 12475--12486, 2021.

\bibitem[Morise et~al.(2017)]{morise2017harvest}
Masanori Morise et~al.
\newblock Harvest: A high-performance fundamental frequency estimator from
  speech signals.
\newblock In \emph{Conference of the International Speech Communication
  Association (INTERSPEECH)}, pages 2321--2325, 2017.

\bibitem[Mottaghi et~al.(2017)Mottaghi, Schenck, Fox, and
  Farhadi]{liquid_perception7_Mottaghi2017SeeTG}
Roozbeh Mottaghi, Connor Schenck, Dieter Fox, and Ali Farhadi.
\newblock See the glass half full: Reasoning about liquid containers, their
  volume and content.
\newblock \emph{International Conference on Computer Vision (ICCV)}, pages
  1889--1898, 2017.
\newblock URL \url{https://api.semanticscholar.org/CorpusID:7410030}.

\bibitem[Narasimhan et~al.(2022)Narasimhan, Zhang, Eisner, Lin, and
  Held]{liquid_perception1_Narasimhan2022SelfsupervisedTL}
Gautham~Narayan Narasimhan, Kai Zhang, Ben Eisner, Xingyu Lin, and David Held.
\newblock Self-supervised transparent liquid segmentation for robotic pouring.
\newblock \emph{IEEE International Conference on Robotics and Automation
  (ICRA)}, pages 4555--4561, 2022.
\newblock URL \url{https://api.semanticscholar.org/CorpusID:247222673}.

\bibitem[Owens and Efros(2018)]{owens2018audio}
Andrew Owens and Alexei~A Efros.
\newblock Audio-visual scene analysis with self-supervised multisensory
  features.
\newblock In \emph{European Conference on Computer Vision (ECCV)}, pages
  631--648, 2018.

\bibitem[Owens et~al.(2016)Owens, Wu, McDermott, Freeman, and
  Torralba]{owens2016ambient}
Andrew Owens, Jiajun Wu, Josh~H McDermott, William~T Freeman, and Antonio
  Torralba.
\newblock Ambient sound provides supervision for visual learning.
\newblock In \emph{European Conference on Computer Vision (ECCV)}, pages
  801--816. Springer, 2016.

\bibitem[Pan et~al.(2016)Pan, Park, and
  Manocha]{pouring2_Pan_Park_Manocha_2016}
Zherong Pan, Chonhyon Park, and Dinesh Manocha.
\newblock Robot motion planning for pouring liquids.
\newblock \emph{Proceedings of the International Conference on Automated
  Planning and Scheduling}, 26\penalty0 (1):\penalty0 518--526, Mar. 2016.
\newblock \doi{10.1609/icaps.v26i1.13787}.
\newblock URL \url{https://ojs.aaai.org/index.php/ICAPS/article/view/13787}.

\bibitem[Peng and Reiss(2018)]{peng2018can}
He~Peng and Joshua~D Reiss.
\newblock Why can you hear a difference between pouring hot and cold water? an
  investigation of temperature dependence in psychoacoustics.
\newblock In \emph{Audio Engineering Society Convention 145}. Audio Engineering
  Society, 2018.

\bibitem[Perfecto et~al.(2019{\natexlab{a}})Perfecto, Donnelly, and
  Critcher]{container_size_doi:10.1177/0956797618813319}
Hannah Perfecto, Kristin Donnelly, and Clayton~R. Critcher.
\newblock Volume estimation through mental simulation.
\newblock \emph{Psychological Science}, 30, 2019{\natexlab{a}}.

\bibitem[Perfecto et~al.(2019{\natexlab{b}})Perfecto, Donnelly, and
  Critcher]{perfecto2019volume}
Hannah Perfecto, Kristin Donnelly, and Clayton~R Critcher.
\newblock Volume estimation through mental simulation.
\newblock \emph{Psychological science}, 30\penalty0 (1):\penalty0 80--91,
  2019{\natexlab{b}}.

\bibitem[Piacenza et~al.(2022)Piacenza, Lee, and Isler]{pouring9_9811898}
Pedro Piacenza, Daewon Lee, and Volkan Isler.
\newblock Pouring by feel: An analysis of tactile and proprioceptive sensing
  for accurate pouring.
\newblock In \emph{IEEE International Conference on Robotics and Automation
  (ICRA)}, pages 10248--10254, 2022.
\newblock \doi{10.1109/ICRA46639.2022.9811898}.

\bibitem[Pykett(2013)]{pykett_organ_pipes}
C.~E. Pykett.
\newblock End corrections, natural frequencies, tone colour and physical
  modelling of organ pipes, 2013.

\bibitem[Rayleigh(1896)]{rayleigh1896theory}
John William Strutt~Baron Rayleigh.
\newblock \emph{The theory of sound}, volume~2.
\newblock Macmillan, 1896.

\bibitem[Riou et~al.(2023)Riou, Lattner, Hadjeres, and Peeters]{riou2023pesto}
Alain Riou, Stefan Lattner, Ga{\"e}tan Hadjeres, and Geoffroy Peeters.
\newblock Pesto: Pitch estimation with self-supervised
  transposition-equivariant objective.
\newblock In \emph{International Society for Music Information Retrieval
  Conference (ISMIR)}, 2023.

\bibitem[Rocchesso et~al.(2003)Rocchesso, Ottaviani, Fontana, and
  Avanzini]{rocchesso2003size}
Davide Rocchesso, Laura Ottaviani, Federico Fontana, and Federico Avanzini.
\newblock Size, shape, and material properties of sound models.
\newblock \emph{The sounding object}, pages 95--110, 2003.

\bibitem[Russell(2023)]{russell2023identifying}
Michael~K Russell.
\newblock Identifying a sound-producing object’s direction of motion and
  change in speed.
\newblock \emph{Auditory Perception \& Cognition}, 6\penalty0 (3-4):\penalty0
  353--368, 2023.

\bibitem[Schenck and
  Fox(2016{\natexlab{a}})]{liquid_perception3_schenck2016detection}
Connor Schenck and Dieter Fox.
\newblock Detection and tracking of liquids with fully convolutional networks,
  2016{\natexlab{a}}.

\bibitem[Schenck and
  Fox(2016{\natexlab{b}})]{liquid_perception4_Schenck2016TowardsLT}
Connor Schenck and Dieter Fox.
\newblock Towards learning to perceive and reason about liquids.
\newblock In \emph{International Symposium on Experimental Robotics},
  2016{\natexlab{b}}.
\newblock URL \url{https://api.semanticscholar.org/CorpusID:12918749}.

\bibitem[Schenck and
  Fox(2017{\natexlab{a}})]{liquid_perception8_Schenck2017PerceivingAR}
Connor Schenck and Dieter Fox.
\newblock Perceiving and reasoning about liquids using fully convolutional
  networks.
\newblock \emph{The International Journal of Robotics Research}, 37:\penalty0
  452 -- 471, 2017{\natexlab{a}}.
\newblock URL \url{https://api.semanticscholar.org/CorpusID:6123383}.

\bibitem[Schenck and Fox(2017{\natexlab{b}})]{pouring1_7989307}
Connor Schenck and Dieter Fox.
\newblock Visual closed-loop control for pouring liquids.
\newblock In \emph{IEEE International Conference on Robotics and Automation
  (ICRA)}, pages 2629--2636, 2017{\natexlab{b}}.
\newblock \doi{10.1109/ICRA.2017.7989307}.

\bibitem[Scott(2017)]{tomscott_youtube}
Tom Scott.
\newblock You can hear the difference between hot and cold water, March 2017.
\newblock URL \url{https://www.youtube.com/watch?v=Ri_4dDvcZeM}.

\bibitem[Sermanet et~al.(2018)Sermanet, Lynch, Chebotar, Hsu, Jang, Schaal,
  Levine, and Brain]{sermanet2018time}
Pierre Sermanet, Corey Lynch, Yevgen Chebotar, Jasmine Hsu, Eric Jang, Stefan
  Schaal, Sergey Levine, and Google Brain.
\newblock Time-contrastive networks: Self-supervised learning from video.
\newblock In \emph{IEEE International Conference on Robotics and Automation
  (ICRA)}, pages 1134--1141. IEEE, 2018.

\bibitem[Shi et~al.(2022)Shi, Hsu, Lakhotia, and Mohamed]{shi2022learning}
Bowen Shi, Wei-Ning Hsu, Kushal Lakhotia, and Abdelrahman Mohamed.
\newblock Learning audio-visual speech representation by masked multimodal
  cluster prediction.
\newblock \emph{arXiv:2201.02184}, 2022.

\bibitem[Traer and McDermott(2018)]{traer2018intuitive}
James Traer and J~McDermott.
\newblock Intuitive physical inference from sound.
\newblock In \emph{2018 Conference on Cognitive Computational Neuroscience},
  pages 2018--1057, 2018.

\bibitem[Van~der Maaten and Hinton(2008)]{van2008visualizing}
Laurens Van~der Maaten and Geoffrey Hinton.
\newblock Visualizing data using t-sne.
\newblock \emph{Journal of Machine Learning Research (JMLR)}, 9\penalty0 (11),
  2008.

\bibitem[Vaswani(2017)]{vaswani2017attention}
A~Vaswani.
\newblock Attention is all you need.
\newblock \emph{Advances in Neural Information Processing Systems (NeurIPS)},
  2017.

\bibitem[Velasco et~al.(2013{\natexlab{a}})Velasco, Jones, King, and
  Spence]{temperature1_https://doi.org/10.1111/joss.12052}
Carlos Velasco, Russ Jones, Scott King, and Charles Spence.
\newblock The sound of temperature: What information do pouring sounds convey
  concerning the temperature of a beverage.
\newblock \emph{Journal of Sensory Studies}, 28, 2013{\natexlab{a}}.

\bibitem[Velasco et~al.(2013{\natexlab{b}})Velasco, Jones, King, and
  Spence]{velasco2013sound}
Carlos Velasco, Russ Jones, Scott King, and Charles Spence.
\newblock The sound of temperature: What information do pouring sounds convey
  concerning the temperature of a beverage.
\newblock \emph{Journal of Sensory Studies}, 28\penalty0 (5):\penalty0
  335--345, 2013{\natexlab{b}}.

\bibitem[Webster and Davies(2010)]{webster2010use}
Emile~S Webster and Clive~E Davies.
\newblock The use of helmholtz resonance for measuring the volume of liquids
  and solids.
\newblock \emph{Sensors}, 10\penalty0 (12):\penalty0 10663--10672, 2010.

\bibitem[Wilson et~al.(2019)Wilson, Sterling, and Lin]{Wilson2019AnalyzingLP}
Justin Wilson, Auston Sterling, and Ming Lin.
\newblock Analyzing liquid pouring sequences via audio-visual neural networks.
\newblock In \emph{IEEE/RSJ International Conference on Intelligent Robots and
  Systems (IROS)}, 2019.

\bibitem[Wu et~al.(2016)Wu, Lim, Zhang, Tenenbaum, and Freeman]{phys101}
Jiajun Wu, Joseph~J Lim, Hongyi Zhang, Joshua~B Tenenbaum, and William~T
  Freeman.
\newblock Physics 101: Learning physical object properties from unlabeled
  videos.
\newblock In \emph{British Machine Vision Conference (BMVC)}, 2016.

\bibitem[Wu et~al.(2018)Wu, Lin, Wang, Hu, Niebles, and Sun]{wu2018liquid}
Tz-Ying Wu, Juan-Ting Lin, Tsun-Hsuang Wang, Chan-Wei Hu, Juan~Carlos Niebles,
  and Min Sun.
\newblock Liquid pouring monitoring via rich sensory inputs, 2018.

\bibitem[Zellers et~al.(2022)Zellers, Lu, Lu, Yu, Zhao, Salehi, Kusupati,
  Hessel, Farhadi, and Choi]{zellers2022merlot}
Rowan Zellers, Jiasen Lu, Ximing Lu, Youngjae Yu, Yanpeng Zhao, Mohammadreza
  Salehi, Aditya Kusupati, Jack Hessel, Ali Farhadi, and Yejin Choi.
\newblock Merlot reserve: Neural script knowledge through vision and language
  and sound.
\newblock In \emph{Conference on Computer Vision and Pattern Recognition
  (CVPR)}, pages 16375--16387, 2022.

\bibitem[Zheng(2021)]{zheng2021pouring}
Qi~Zheng.
\newblock Pouring dynamics estimation using gated recurrent units, 2021.

\bibitem[Zwicker and Fastl(2013)]{zwicker2013psychoacoustics}
Eberhard Zwicker and Hugo Fastl.
\newblock \emph{Psychoacoustics: Facts and models}, volume~22.
\newblock Springer Science \& Business Media, 2013.

\end{thebibliography}
}


\newpage
\appendix
\section{Appendix / supplemental material}

\subsection{Dataset}

\paragraph{Recording setup.} We use the OnePlus Nord CE 5G phone with its in-built microphone to record videos of liquid pouring. The phone is placed against a wall and the video is recorded with the front camera of 16 MP. The container is placed such that it is entirely visible in the camera. The recording sequence is as follows: (i) start recording, (ii) start pouring and pour till the container is full, (iii) stop recording. The recording has few seconds of noise during steps (i) and (ii). We remove this noise and only consider the pouring action from start to end as described below. The videos are of resolution  $1080 \times 1920$ with 30 $\operatorname{FPS}$. The audio is sampled at \SI{44.1}{\kilo\hertz}. For each container, we also record physical measurements such as the base radius, net height, top radius, \etc. These are done manually using a transparent ruler. The container shape, material, type of liquid are all noted before recording.

\paragraph{Pre-processing details.} We resize the videos to $270 \times 480$ and resample the audio at \SI{16}{\kilo\hertz}. We obtain segmentation (and a bounding box) of the container by processing the first frame of the video with LangSAM~\cite{langsam}. A simple text prompt shown below works well across all videos.
\tcbset{
    colback=gray!7,    
    colframe=gray!90,   
    boxrule=0.4mm,      
    width=\linewidth,   
    arc=2mm,            
    boxsep=5pt,         
}
\begin{tcolorbox}
    \textit{liquid container or vessel or a glass or a cup or a bowl kept on a kitchen table}
\end{tcolorbox}

To avoid noise/silence at the start and end of the pouring sound, we annotate the precise start and end time for a subset of 150 videos. We use this to train a per-frame binary classifier with DINO features~\cite{caron2021emerging}. The classifier labels 0 to a frame with empty container and 1 to a filled container.
A score in $(0, 1)$ roughly indicates the fraction of the container filled.
The trained model is used to infer start and end times of pouring for the rest of the videos. The model annotations are manually verified for correctness and manually corrected if needed.

\subsection{Pre-training audio network}
\label{appendix-subsec:audio-network}

\begin{figure}[h]
    \centering
    \includegraphics[width=\linewidth]{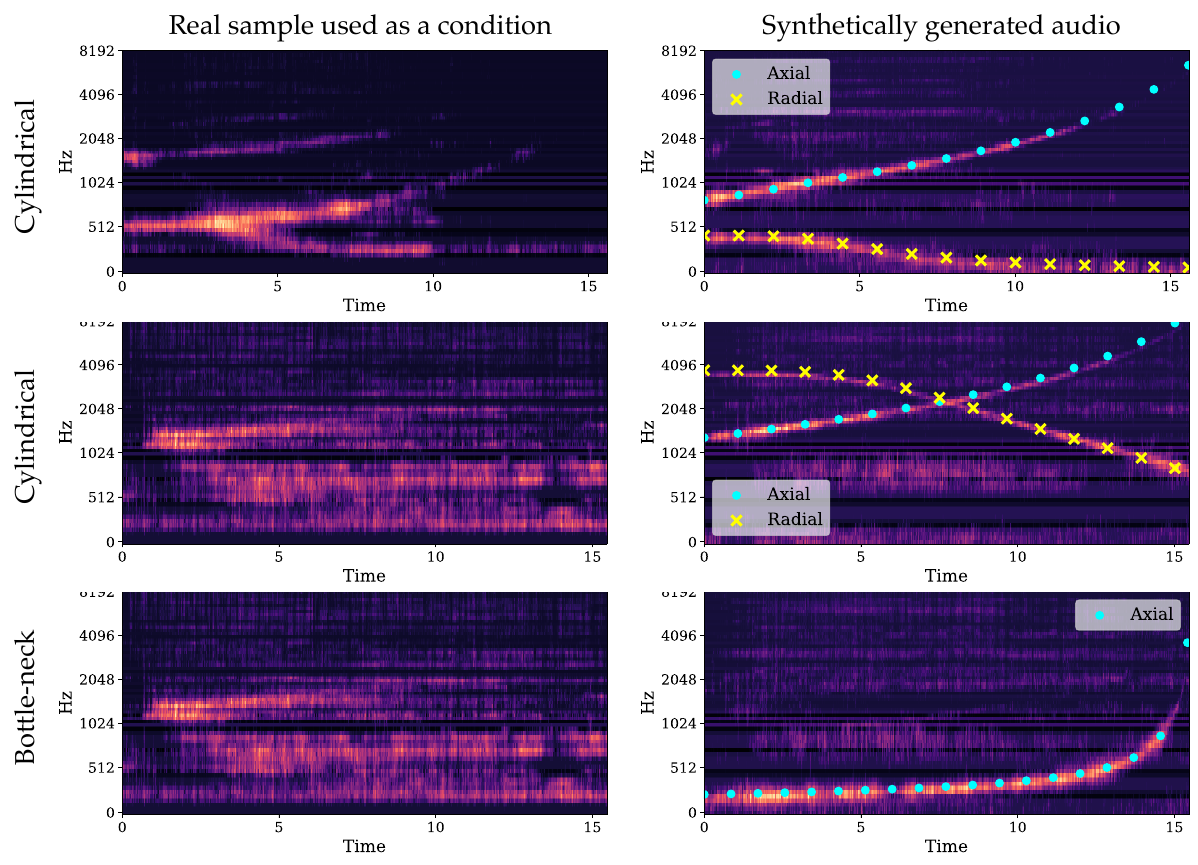}
    \caption{\textbf{Examples of synthetic sounds of pouring.}
        Each synthetic sample is conditioned on a real sample and a random pitch profile.
        The cyan curve is the fundamental of axial resonance and the yellow curve is that of radial resonance.
        Note that while we only use axial resonance in training, our synthetic data pipeline is also capable of generating sounds with radial resonance. The first two rows show sounds generated for cylindrical container while the last row shows that for a bottle-neck container following~\cite{Helmholtz2005OnTS}.
    }
    \label{fig:synthetic-samples}
\end{figure}

\paragraph{Synthetic data.} Some examples of generated sounds are shown in \cref{fig:synthetic-samples} (Right). To generate a single sample, we condition its loudness and residual (background noise) on a real sample from the train set. Given a real sample and a randomly generated pitch profile $f(t), \forall t$, we generate a simulated sample with the desired pitch profile. The first two rows in \cref{fig:synthetic-samples} show generated sound of pouring in cylindrical containers.
The cyan curve is the fundamental of axial resonance and the yellow curve is that of radial resonance.
Note that while we only use axial resonance in training, our synthetic data pipeline is also capable of generating sounds with radial resonance.
For bottle-neck containers, following \cite{Helmholtz2005OnTS}, the pitch follows a different equation. Our synthetic data pipeline is easily adaptable to such bottle-neck containers as well (last row in \cref{fig:synthetic-samples}). In this case, only axial resonance is observed.

\paragraph{Architecture and training details.}
The audio network architecture is based on \texttt{wav2vec2}~\cite{baevski2020wav2vec}.
It takes in raw audio samples $\mathbf{x} \in \mathbb{R}^{N}$ and outputs a distribution over wavelengths $\mathbf{y} \in \mathbb{R}^{T \times K}$ where $T$ is the number of time frames and $K$ is the number of wavelength bins.

The network consists of (i) a feature encoder that converts raw samples to feature vectors and (ii) a transformer encoder to capture information from the entire sequence.
The feature encoder is a 1D CNN: $\mathbb{R}^{N} \rightarrow \mathbb{R}^{T \times d}$ where $T$ is the number of frames.  The CNN encoder processes each frame of $400$ samples with a hop length of $320$. If the audio is \SI{1}{\second} long, then $T = 49$ frames. It consists of seven blocks and the temporal convolutions in each block have 512
channels with strides (5,2,2,2,2,2,2) and kernel widths (10,3,3,3,3,2,2).
Each block has a layer normalization and a GELU activation function~\cite{hendrycks2016gaussian}.
We use the BASE Transformer variant~\cite{vaswani2017attention} that has 12 transformer blocks, model dimension 768, inner dimension (FFN) 3,072 and 8 attention heads. The original model in \cite{baevski2020wav2vec} uses a CNN for relative position embeddings. In our use case, it is necessary to strengthen absolute position information, since we want to model wavelength as a function of time. Thus, we also use absolute position encoding using sinusoids as in \cite{vaswani2017attention}.

\subsection{Pre-training visual network}
\label{appendix-subsec:visual-network}

To co-supervise the audio network, we first train a visual network to detect the length of the air column $l(t)$ and the radius of the container $R$.

\begin{figure}
    \centering
    \includegraphics[width=\linewidth]{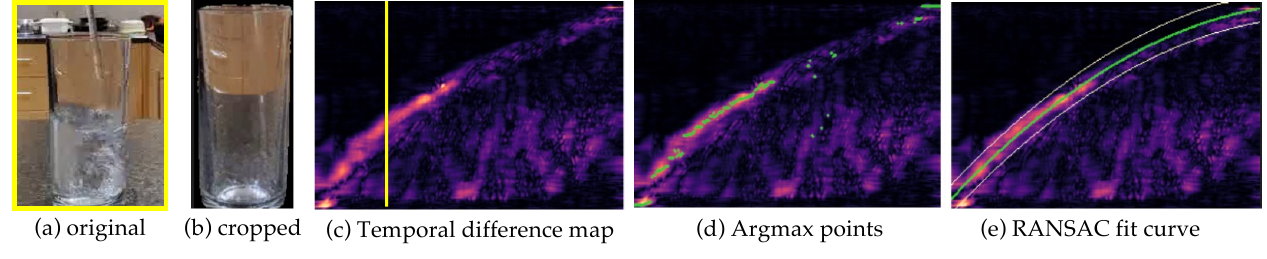}
    \captionsetup{skip=1mm}
    \caption{\textbf{Obtaining pseudo labels for visual pre-training.} For a given video (sample frame at time $t$ in (a)), we mask out the container (b) and then compute pixel wise differences which leads to a tensor of size $F \times H \times W$.
        We average across width and get a temporal difference map (c). We fit a polynomial with RANSAC (e) to the argmax points (d) which gives us an estimate for $l(t), \forall t$.}
    \label{fig:tdm}
\end{figure}

\paragraph{Obtaining pseudo ground truths.} Consider a video with $F$ frames of liquid pouring. Since the camera and the container are static, the temporal difference between consecutive frames serves as a strong signal for the liquid height at a given time. First, for each frame, we black out pixels outside the container using the segmentation map.
Then, we compute \textit{temporal difference maps} simply by computing the normalized pixel-wise difference between consecutive frames.
We take the mean over image width which gives as a single map of size $F \times H$ where $H$ is the image height.
These are Gaussian smoothened over $F$ and $H$ dimensions.
An example is shown in \cref{fig:tdm} (c). Then, for each frame, we pick the highest intensity ordinate shown as green scatter points in \cref{fig:tdm} (d). To avoid noise, we fit a second-order polynomial with RANSAC to these points as shown in \cref{fig:tdm} (e). The container height minus this fit curve gives us a ground truth estimate of $l(t)$. These are then used to train a DINO based video network. We train a deep network with these ground truths because it can generalize better to unseen containers and other variations over the handcrafted method.

\begin{figure}[h]
    \centering
    \includegraphics[width=\textwidth]{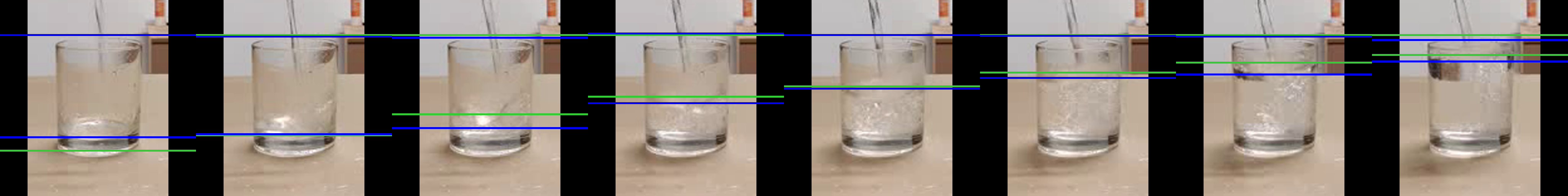}
    \caption{\textbf{Qualitative result for visual pre-training.} Video model predictions (blue) of height of liquid with pseudo-ground truth (green) on an example from the test set I.}
    \label{fig:visual-pretraining-result-v1}
\end{figure}

\paragraph{Architecture and training details.} Given the strong dense visual understanding of DINO~\cite{caron2021emerging}, we use it as an image backbone and adapt it to detect length of air column in videos. For each frame, we extract the CLS token features out of DINO. The sequence of CLS tokens is projected to a lower dimension and passed through to a light Transformer to contextualize features. Sinusoidal position encodings are added.
The Transformer has a single layer with 4 heads and 128 dimensions.
Then, at each time step, an MLP regressor head ($128 \rightarrow 64 \rightarrow 64 \rightarrow 2$) regresses a 1D bounding box that denotes the top and bottom of the air column which is used to calculate the length $\hat{l}(t)$. The model is trained with MSE loss. An example qualitative result is shown in \cref{fig:visual-pretraining-result-v1}. The green lines denote ground truths obtained from the previous step, and the blue lines denote the model predictions.

\begin{figure}[h]
    \centering
    \includegraphics[width=\textwidth]{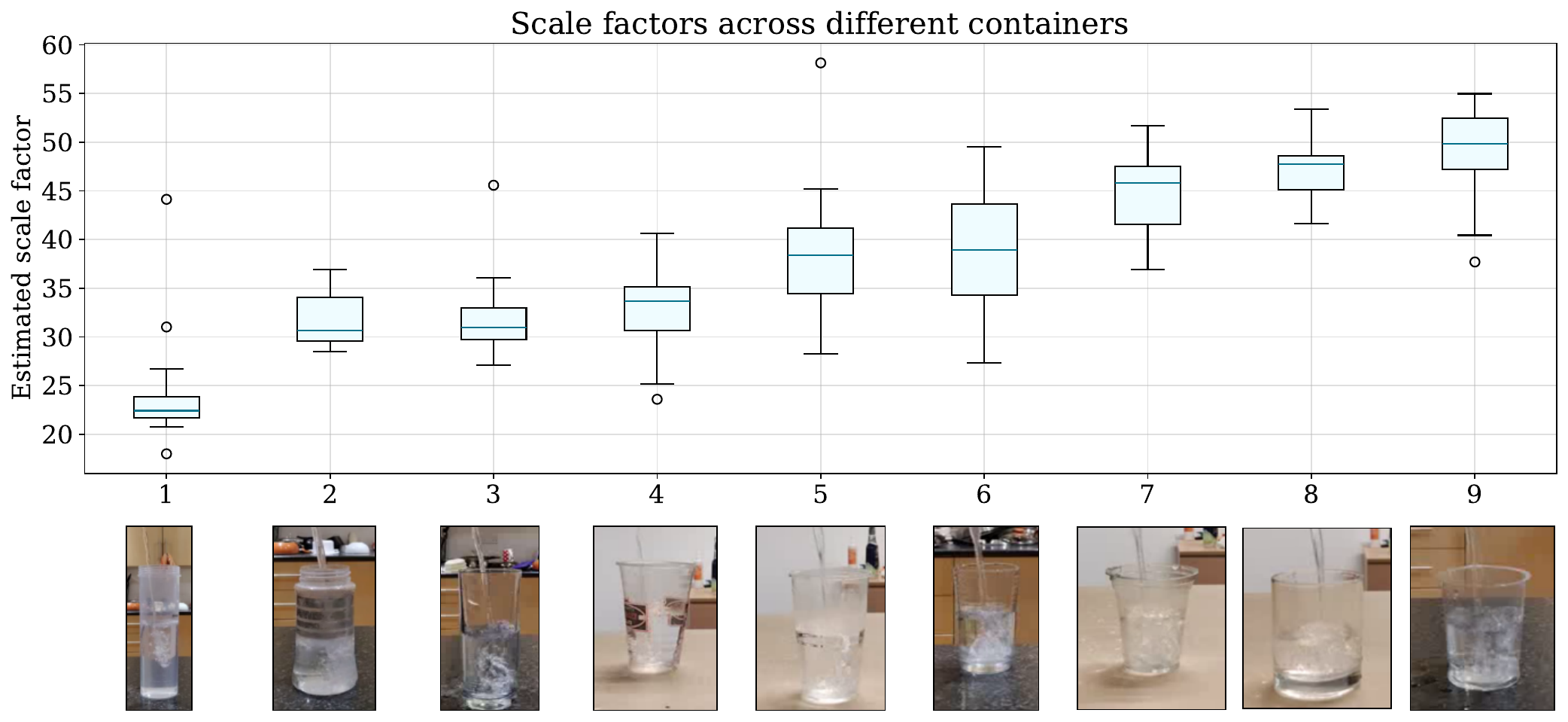}
    \caption{\textbf{Estimated scale factors for a subset of containers.} Generally, larger containers (\eg, containers 1 and 4) have smaller scale factors since the camera needs to be placed further apart to produce an image where the container is roughly at the center.
        This is indeed the case with our scale factors computed as the ratio of visual outputs to audio outputs.
    }
    \label{fig:example-scale-factors}
\end{figure}

\paragraph{Verifying scale factor computation.} As shown in \cref{eq:alpha}, the theoretical estimate of scale factors is given by
\[
    \alpha := \frac{1}{Z}\frac{f}{s}.
\]
Assuming generic values of the intrinsic parameters $f, s$ and depth $Z$ varyng between $10$-\SI{50}{\centi\metre}, we should get $\alpha \in [30, 80]$. Our empirical estimates match this range of values. Furthermore, since $\alpha$ is inversely related to depth, and container size is directly related to the depth (larger container needs to be kept further away from the camera), $\alpha$ should be inversely related to container size. This is verified in our estimates as shown in \cref{fig:example-scale-factors}. Videos with a large container (\eg, container $1$) have smaller scale factors compared to those with a small container (\eg, container $9$).

\subsection{Other results}
\label{appendix-subsec:other-results}

\paragraph{Full results}
We present a more comprehensive table with all the results (including those presented in \cref{tab:pitch-detection-results} and \cref{tab:physical-properties} in the main paper) in \cref{tab:main-results}. Methods that do well in estimating the length of air column generally tend to do well in estimating all other properties. This suggests the importance of accurate estimation of pitch (and thus, length of air column) merely from the sound of pouring.

\begin{table}[h]
    \centering
    \resizebox{\columnwidth}{!}{%
        \begin{tabular}{lccccccc}
            \toprule
                                                                        & \multicolumn{1}{c}{\textbf{Length}} & \multicolumn{2}{c}{\textbf{Static properties}} & \multicolumn{4}{c}{\textbf{Dynamic properties}}                                                                                                                     \\ \smallskip
                                                                        & \textbf{of air column} $\downarrow$ & Radius $\downarrow$                            & Height $\downarrow$                             & Flow rate $\downarrow$ & \multicolumn{3}{c}{Time to fill \small (s) $\downarrow$}                                 \\ \smallskip
                                                                        & $l(t)$ \small (cm)                  & $H$ \small (cm)                                & $R$ \small (cm)                                 & $Q(t)$ \small (ml/s)   & \small 25\%                                              & \small 50\%   & \small 75\%   \\
            \midrule
            \textcolor{blue}{\textit{Test set I (seen containers)}}     &                                     &                                                &                                                 &                        &                                                          &               &               \\[1mm]
            \multicolumn{8}{l}{\cellcolor[HTML]{f2f2f2}\textit{\textbf{Baselines}}}                                                                                                                                                                                                                                                  \\[1mm]
            Spectrogram argmax                                          & 4.60                                & 5.52                                           & 13.2                                            & 54.7                   & 7.98                                                     & 7.35          & 7.99          \\
            CREPE~\cite{kim2018crepe}                                   & 7.61                                & 9.00                                           & 6.75                                            & 307.7                  & 9.61                                                     & 6.08          & 7.00          \\
            PESTO~\cite{riou2023pesto}                                  & 11.7                                & 8.85                                           & 6.77                                            & 339.7                  & 7.95                                                     & 6.92          & 7.80          \\
            Yin~\cite{de2002yin}                                        & 30.8                                & 10.9                                           & 6.77                                            & 1447.8                 & 7.17                                                     & 6.40          & 6.56          \\
            \multicolumn{8}{l}{\cellcolor[HTML]{f2f2f2}\textit{\textbf{Ours}}}                                                                                                                                                                                                                                                       \\[1mm]
            Audio-only                                                  & 0.78                                & {\textbf{2.23}}                                & 1.62                                            & 25.2                   & \textbf{3.96}                                            & 1.62          & 1.53          \\
            Co-supervised                                               & {\textbf{0.60}}                     & 2.27                                           & \textbf{1.39}                                   & \textbf{22.5}          & 4.16                                                     & \textbf{1.49} & \textbf{1.07} \\
            \midrule
            \textcolor{blue}{\textit{ Test set II (unseen containers)}} &                                     &                                                &                                                 &                        &                                                          &               &               \\[1mm]
            \multicolumn{8}{l}{\cellcolor[HTML]{f2f2f2}\textit{\textbf{Baselines}}}                                                                                                                                                                                                                                                  \\[1mm]
            Spectrogram argmax                                          & 5.11                                & 5.89                                           & 12.2                                            & 65.2                   & 8.69                                                     & 8.51          & 8.26          \\
            CREPE~\cite{kim2018crepe}                                   & 9.39                                & 9.15                                           & 6.21                                            & 403.41                 & 10.5                                                     & 6.62          & 5.80          \\
            PESTO~\cite{riou2023pesto}                                  & 10.55                               & 9.06                                           & 6.23                                            & 259.21                 & 7.85                                                     & 8.65          & 7.36          \\
            Yin~\cite{de2002yin}                                        & 27.28                               & 10.87                                          & 6.23                                            & 1095.12                & 7.96                                                     & 7.00          & 8.64          \\
            \multicolumn{8}{l}{\cellcolor[HTML]{f2f2f2}\textit{\textbf{Ours}}}                                                                                                                                                                                                                                                       \\[1mm]
            Audio-only                                                  & 0.82                                & \textbf{2.77}                                  & 2.24                                            & 45.7                   & 4.39                                                     & 3.44          & 2.66          \\
            Co-supervised                                               & \textbf{0.71}                       & 2.85                                           & \textbf{1.88}                                   & \textbf{40.4}          & \textbf{4.10}                                            & \textbf{2.99} & \textbf{2.21} \\
            \bottomrule
        \end{tabular}%
    }
    \caption{
        \textbf{Full quantitative results on the evaluation sets.}
        Here, $l(t)$ is the length of air column, $H$ is the container
        height, $R$ is the container radius, $Q(t)$ denotes the
        volume flow rate of pouring and $\tau(t)$ denotes the time
        to fill 50\% of the container. On both test sets, an easier
        set I consisting of seen containers and a harder set II of
        unseen (opaque) containers, the audio model co-supervised
        with a video teacher outperforms the audio-only model only
        pre-trained on synthetic samples. Methods that do well
        in estimating the length of air column generally tend to
        do well in estimating all other properties.
    }
    \label{tab:main-results}
\end{table}

\begin{figure}[h]
    \centering
    \includegraphics[width=\linewidth]{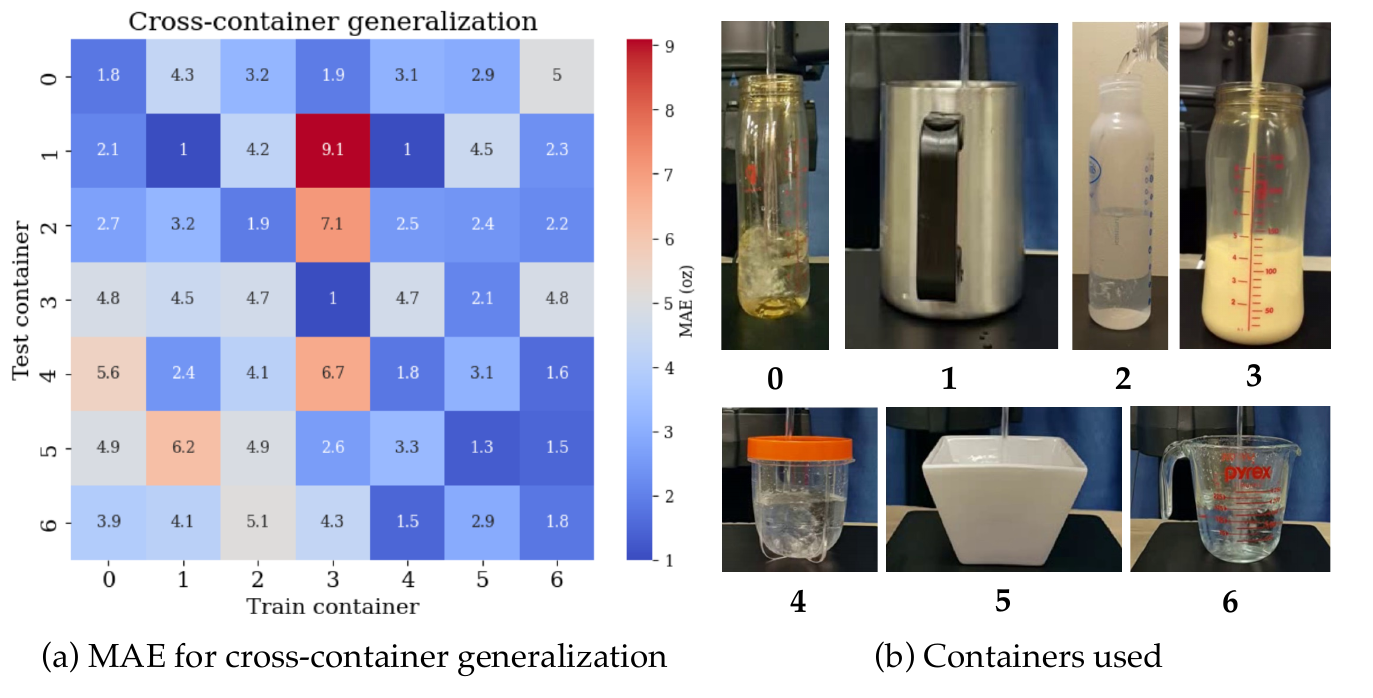}
    \caption{\textbf{Cross-container generalization in liquid mass estimation.}
        On the dataset by \citet{Wilson2019AnalyzingLP}, on the seven container-liquid cases, we train to detect liquid mass on pouring sounds of a single container and test on those of every container.
        We find promising generalization except while generalizing from milk pouring to water pouring.
        This is likely due to their difference densities.
        Larger containers and smaller containers generalize well amongst themselves but not across each other.}
    \label{fig:cross-container}
\end{figure}

\paragraph{Cross-container generalization}
As highlighted in \cite{Wilson2019AnalyzingLP}, a key shortcoming of fully supervised models for liquid mass estimation from sound of pouring is that they do not generalize across containers. We test this with our co-supervised model and find promising generalization to new containers as reported in \cref{fig:cross-container}. On the dataset by \citet{Wilson2019AnalyzingLP}, on the seven container-liquid cases, we train on a single container and test on every container. We find promising generalization except while generalizing from milk pouring to water pouring.
This could be due to different densities of milk and water since mass estimation needs both density and volume.
\[
    m(t) = \rho V(t)
\]
Furthermore, we also find that larger and smaller containers generalize well amonst themselves but not across each other.

\end{document}